\documentclass[letterpaper, 10 pt, conference]{ieeeconf}

\IEEEoverridecommandlockouts

\usepackage[utf8]{inputenc} % allow utf-8 input
\usepackage[T1]{fontenc}    % use 8-bit T1 fonts
\usepackage{hyperref}       % hyperlinks
\usepackage{url}            % simple URL typesetting
\usepackage{booktabs}       % professional-quality tables
\usepackage{amsfonts}       % blackboard math symbols
\usepackage{nicefrac}       % compact symbols for 1/2, etc.
\usepackage{microtype}      % microtypography

\usepackage{times}
\usepackage{color}
\usepackage{pifont}
\usepackage{xspace}

\usepackage{dblfloatfix}
\usepackage{graphicx} % more modern
\usepackage{caption}
\usepackage[font=footnotesize]{caption,subcaption}
\usepackage{wrapfig}
\usepackage{rotating}

\usepackage{algorithm}
\usepackage{algorithmic}

\usepackage{hyperref}

\usepackage{amsmath}
\usepackage{amssymb}

\usepackage{blindtext}

\newcommand{\bo}{\mathbf{o}}
\newcommand{\ba}{\mathbf{a}}
\newcommand{\bd}{\mathbf{d}}
\newcommand{\bg}{\mathbf{g}}
\newcommand{\dx}{\Delta \mathbf{x}}
\newcommand{\loss}{\mathcal{L}}
\newcommand{\lossCE}{\loss^{\textsc{CE}}}
\newcommand{\cost}{\mathcal{C}}
\newcommand{\dataset}{\mathcal{D}}

\usepackage{multirow}
\usepackage{tabularx}
\newcommand{\tabledefaultcolspacing}{6pt}
\newcommand{\tabledefaultrowspacing}{1.2}
\renewcommand{\arraystretch}{\tabledefaultrowspacing}
\newcommand{\specialcell}[2][c]{%
  \begin{tabular}[#1]{@{}c@{}}#2\end{tabular}}

\newcolumntype{Y}{>{\centering\arraybackslash}X}

\newcommand{\ourmethod}{{LaND}\xspace}

\title{\LARGE \bf
\ourmethod: Learning to Navigate from Disengagements
}

\author{\authorblockN{Gregory Kahn, Pieter Abbeel, Sergey Levine}
\authorblockA{Berkeley AI Research (BAIR), University of California, Berkeley}}

\begin{document}

\setlength{\tabcolsep}{2.5pt}
\renewcommand{\arraystretch}{1}
\twocolumn[{%
\renewcommand\twocolumn[1][]{#1}%
\maketitle
\thispagestyle{empty}
\pagestyle{empty}
\vspace*{-25pt}
\begin{center}
    \centering
    % left bottom right up
	\begin{tabularx}{\textwidth}{c *{5}{Y} }

    \includegraphics[height=0.112\textheight]{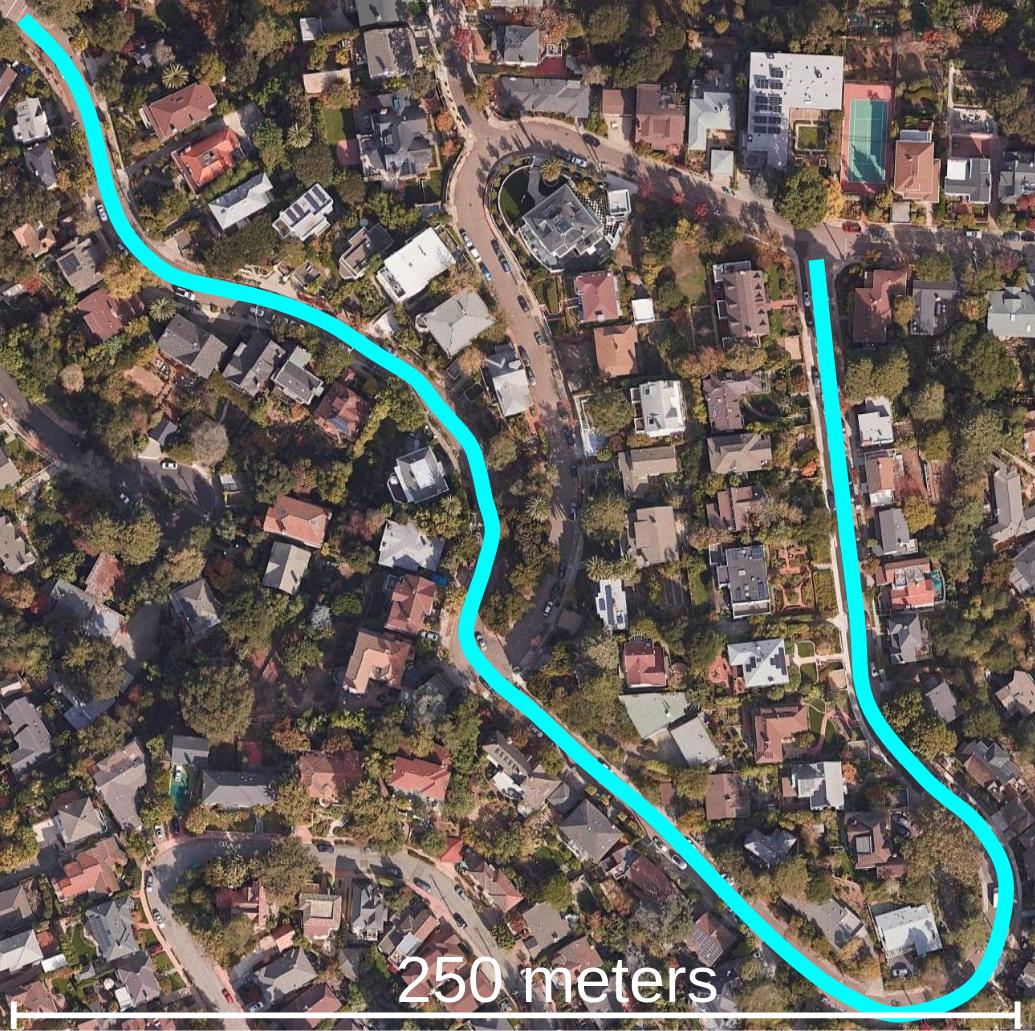} &
    \includegraphics[height=0.112\textheight]{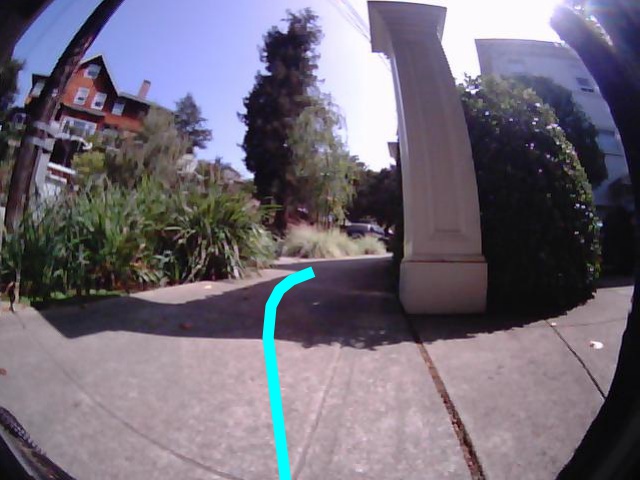} &
    \includegraphics[height=0.112\textheight]{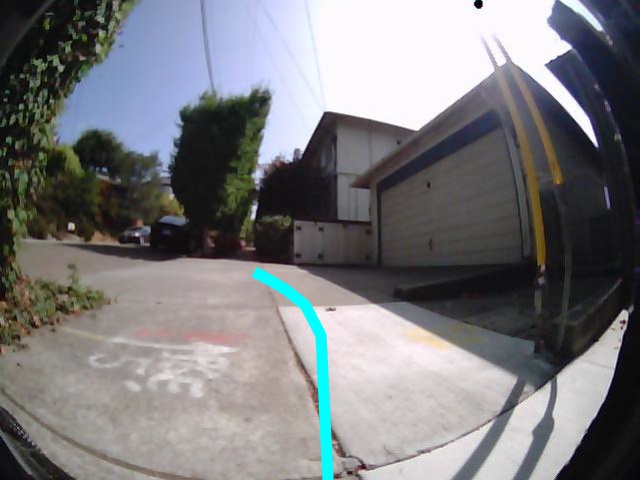} &
    \includegraphics[height=0.112\textheight]{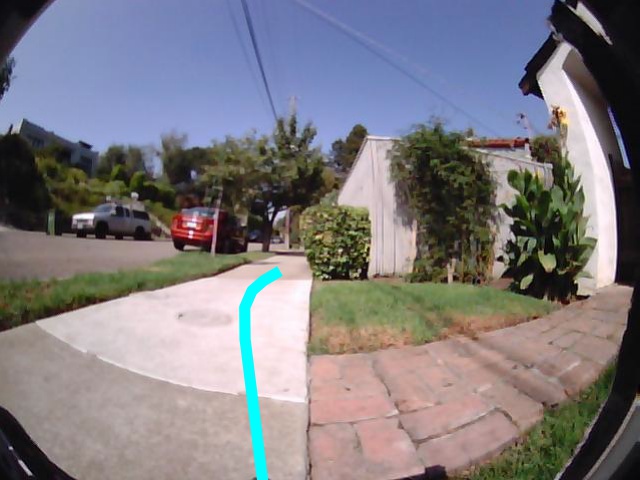} &
    \includegraphics[height=0.112\textheight]{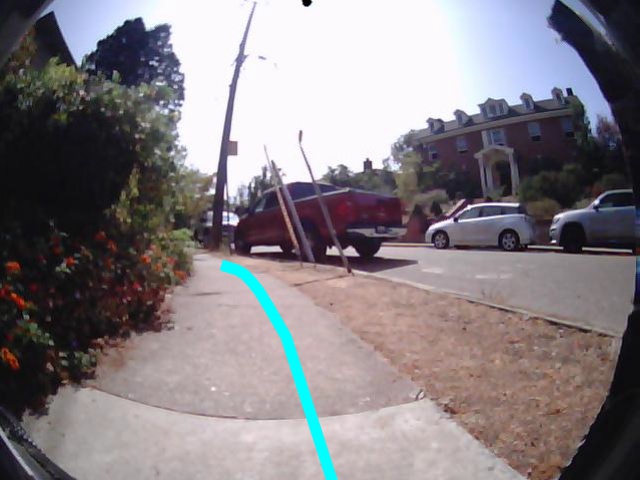} \\

    \includegraphics[height=0.112\textheight]{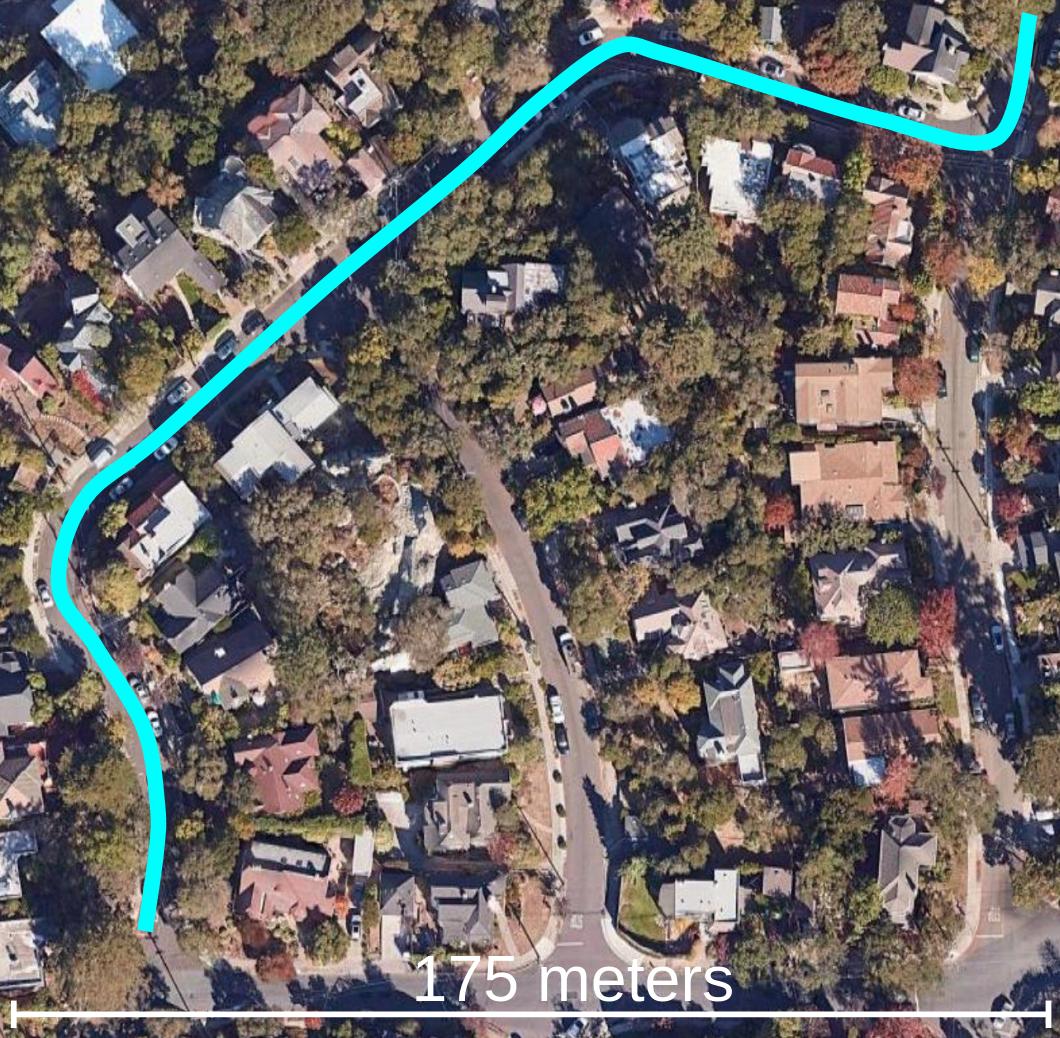} &
    \includegraphics[height=0.112\textheight]{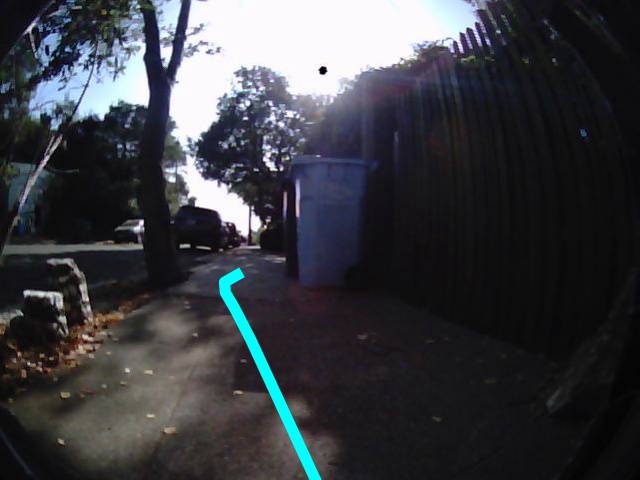} &
    \includegraphics[height=0.112\textheight]{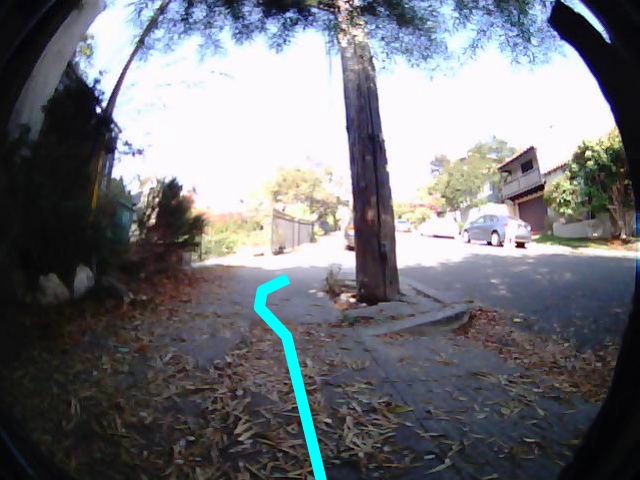} &
    \includegraphics[height=0.112\textheight]{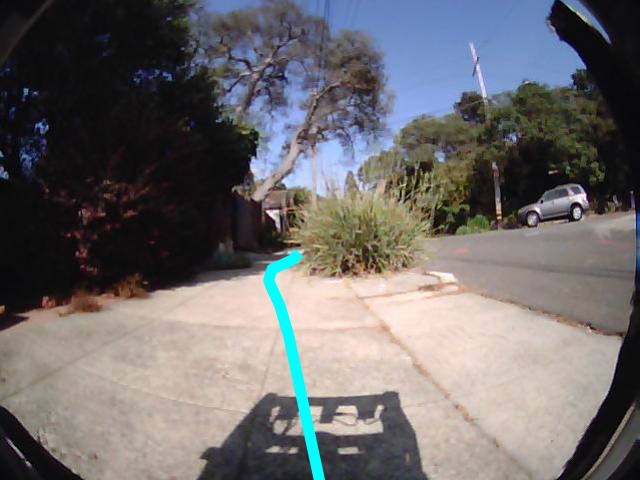} &
    \includegraphics[height=0.112\textheight]{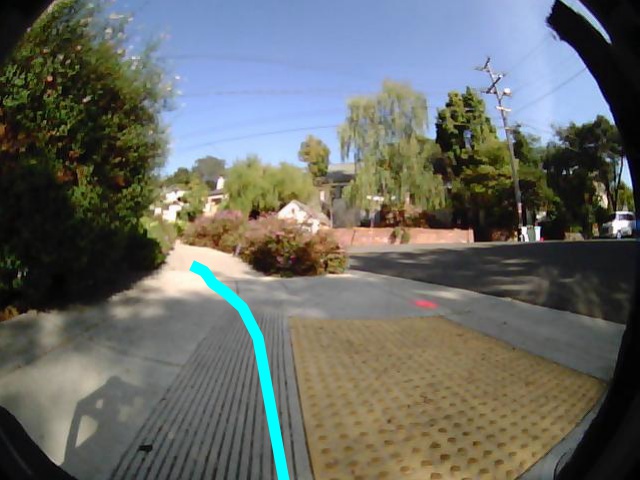} \\
    
    \includegraphics[height=0.112\textheight]{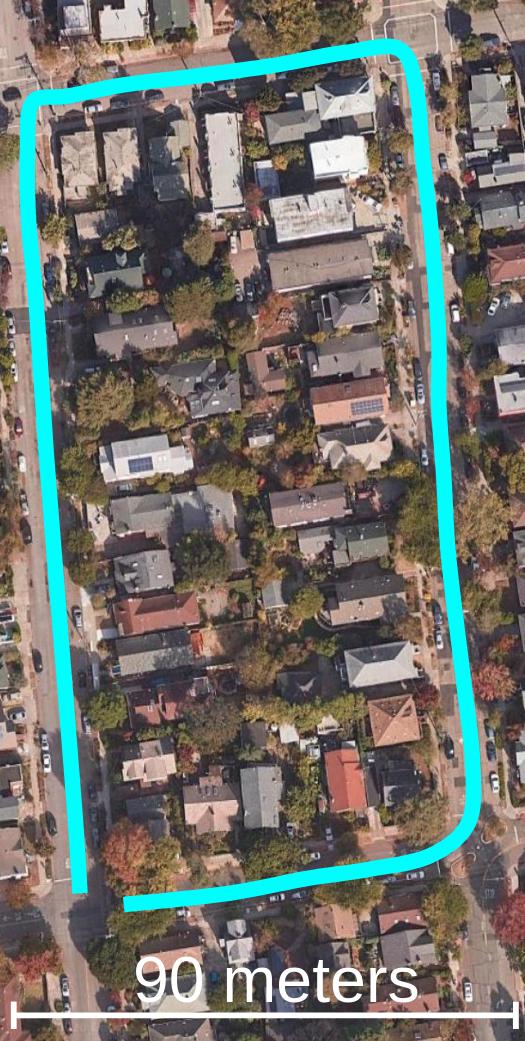} &
    \includegraphics[height=0.112\textheight]{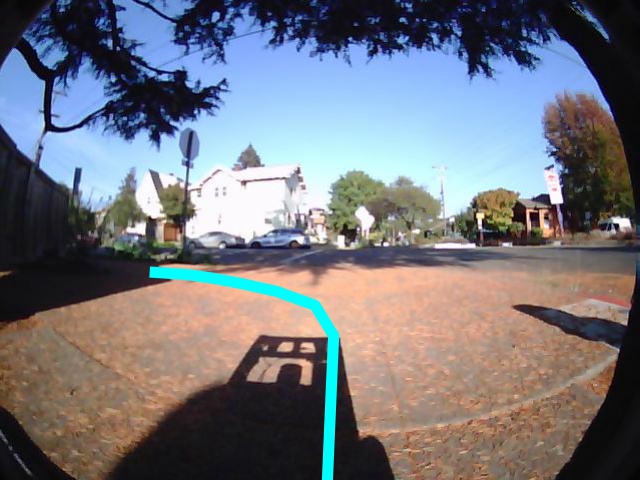} &
    \includegraphics[height=0.112\textheight]{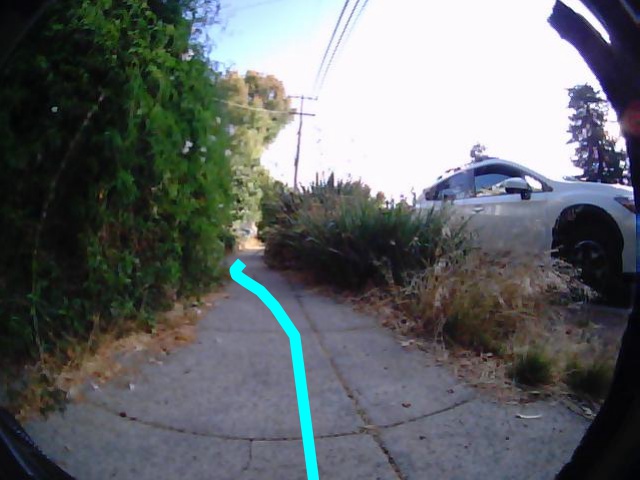} &
    \includegraphics[height=0.112\textheight]{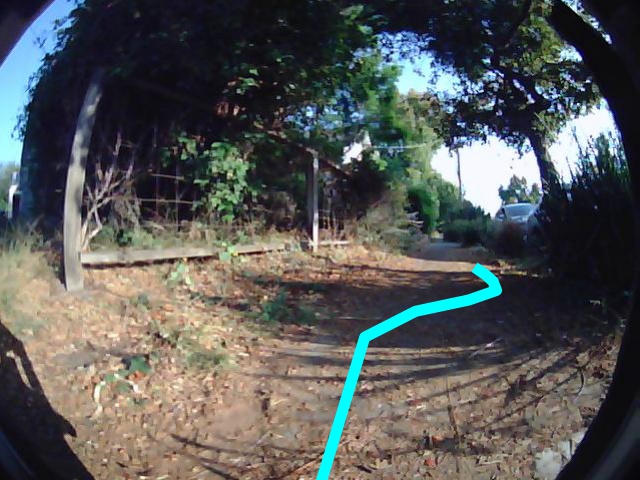} &
    \includegraphics[height=0.112\textheight]{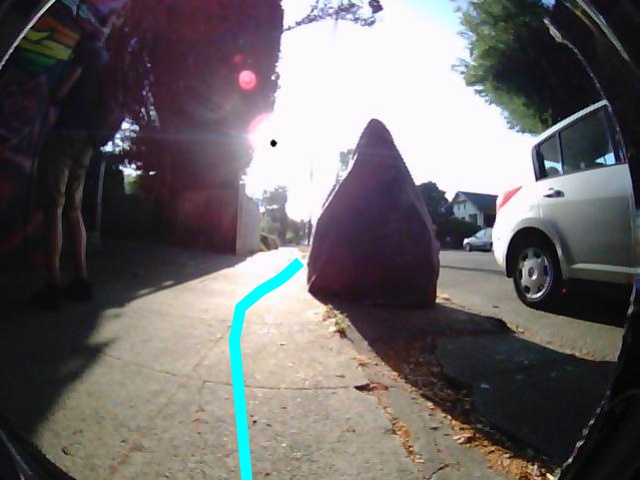}
    
	\end{tabularx}
	\vspace*{-15pt}
	\captionof{figure}{\ourmethod is a learning-based approach for autonomous mobile robot navigation that directly learns from disengagements---any time a human monitor disengages the robot's autonomy. These disengagement datasets are ubiquitous because they are naturally collected during the process of testing these autonomous systems. \ourmethod is able to navigate in a diverse set of sidewalk environments, including parked bicycles, dense foliage, parked cars, sun glare, sharp turns, and unexpected obstacles.}
    \label{fig:teaser}
\end{center}%
}]
\setlength{\tabcolsep}{\tabledefaultcolspacing}
\renewcommand{\arraystretch}{\tabledefaultrowspacing}

%%%%%%%%%%%%%%%%%%%%%%%%%%%%%%%%%%%%%%%%%%%%%%%%%%%%%%%%%%%%%%%%%%%%%%%%%%%%%%%%
\begin{abstract}
Consistently testing autonomous mobile robots in real world scenarios is a necessary aspect of developing autonomous navigation systems. Each time the human safety monitor disengages the robot's autonomy system due to the robot performing an undesirable maneuver, the autonomy developers gain insight into how to improve the autonomy system. However, we believe that these disengagements not only show where the system fails, which is useful for troubleshooting, but also provide a direct learning signal by which the robot can learn to navigate. We present a reinforcement learning approach for learning to navigate from disengagements, or \ourmethod. \ourmethod learns a neural network model that predicts which actions lead to disengagements given the current sensory observation, and then at test time plans and executes actions that avoid disengagements. Our results demonstrate \ourmethod can successfully learn to navigate in diverse, real world sidewalk environments, outperforming both imitation learning and reinforcement learning approaches. Videos, code, and other material are available on our website \url{https://sites.google.com/view/sidewalk-learning}
\end{abstract}

%%%%%%%%%%%%%%%%%%%%%%%%%%%%%%%%%%%%%%%%%%%%%%%%%%%%%%%%%%%%%%%%%%%%%%%%%%%%%%%%
\section{Introduction}

Recent technological advancements have spurred the development of autonomous mobile robots, from sidewalk delivery robots and agricultural inspection quadcopters to autonomous vehicles. One of the primary metrics for measuring progress has been the notion of average distance travelled before disengagement: how far can the robot travel before the robot fails and a human must intervene? Although using these disengagement numbers as a metric for comparing progress is contentious~\cite{DisengagementReport2020}, the general consensus is clear: the better the autonomous system, the less disengagements. However, we believe a shift in perspective is needed. Not only do these disengagements show where the existing system fails, which is useful for troubleshooting, but also that these disengagements provide a \emph{direct learning signal} by which the robot can learn how to navigate. With this perspective, we believe disengagement data is severely underutilized, and in this work we investigate how to learn to navigate using disengagements as a reinforcement signal.

Disengagements are typically used as a tool to debug and improve autonomous mobile robots: run the robot, discover the failure modes, and then rectify the system such that those failure modes are removed. While the first two steps are fairly similar for most developers, the last step---figuring out how to fix the autonomy failure modes---is highly nontrivial and system dependent. This is especially true for the learning-based modules, which are a key component of modern autonomous mobile robots. Improving these learning-based components is a complex process, which could involve designing neural network architectures, hyperparameter tuning, and data collection and labelling; this is a time-consuming, expensive, and technically challenging endeavor that has to be done largely through trial and error.

In this paper, we propose a method for learning to navigate from disengagements, or \ourmethod, which sidesteps this laborious process by directly learning from disengagements. Our key insight is that if the robot can successfully learn to execute actions that avoid disengagement, then the robot will successfully perform the desired task. Crucially, unlike conventional reinforcement learning algorithms, which use task-specific reward functions, our approach does not even need to \emph{know} the task -- the task is specified implicitly through the disengagement signal. However, similar to standard reinforcement learning algorithms, our approach continuously improves because our learning algorithm reinforces actions that avoid disengagements.

Our approach works by leveraging a ubiquitous dataset: the robot's sensory observations (e.g., camera images), commanded actions (e.g., steering angle), and whether the robot autonomy mode was engaged or disengaged. Using this dataset, we then learn a predictive model that takes as input the current observation and a sequence of future commanded actions, and predicts whether the robot will be engaged or disengaged in the future. At test time, we can then use this predictive model to plan and execute actions that avoid disengagement while navigating towards a desired goal location.

Our work has three primary contributions. First, we propose that disengagements provide a strong, direct supervision signal for autonomous mobile robots, which is already being collected in many commonly used real-world pipelines. Second, we introduce our learning to navigate from disengagements algorithm, or \ourmethod, which is a simple, effective, and scalable framework for creating autonomous mobile robots. Third, we demonstrate our approach on a real world ground robot in diverse and complex sidewalk navigation environments (Fig.~\ref{fig:teaser}), and show our method outperforms state-of-the-art imitation learning and reinforcement learning approaches.

\setcounter{figure}{3}
\setlength{\tabcolsep}{2.5pt}
\renewcommand{\arraystretch}{1}
\begin{figure*}[b]
    \vspace*{-10pt}
	\begin{tabularx}{\textwidth}{ll *{4}{Y} }
     & & Autonomous (robot) & Disengagement (human) & Course Correction (human) & Autonomous (robot) \\	
	
	 \multirow{3}{*}{\rotatebox[origin=t]{90}{Types of Disengagements}} &
	 \rotatebox[origin=t]{90}{Collision\hspace*{-50pt}} &
	 \includegraphics[width=0.225\textwidth]{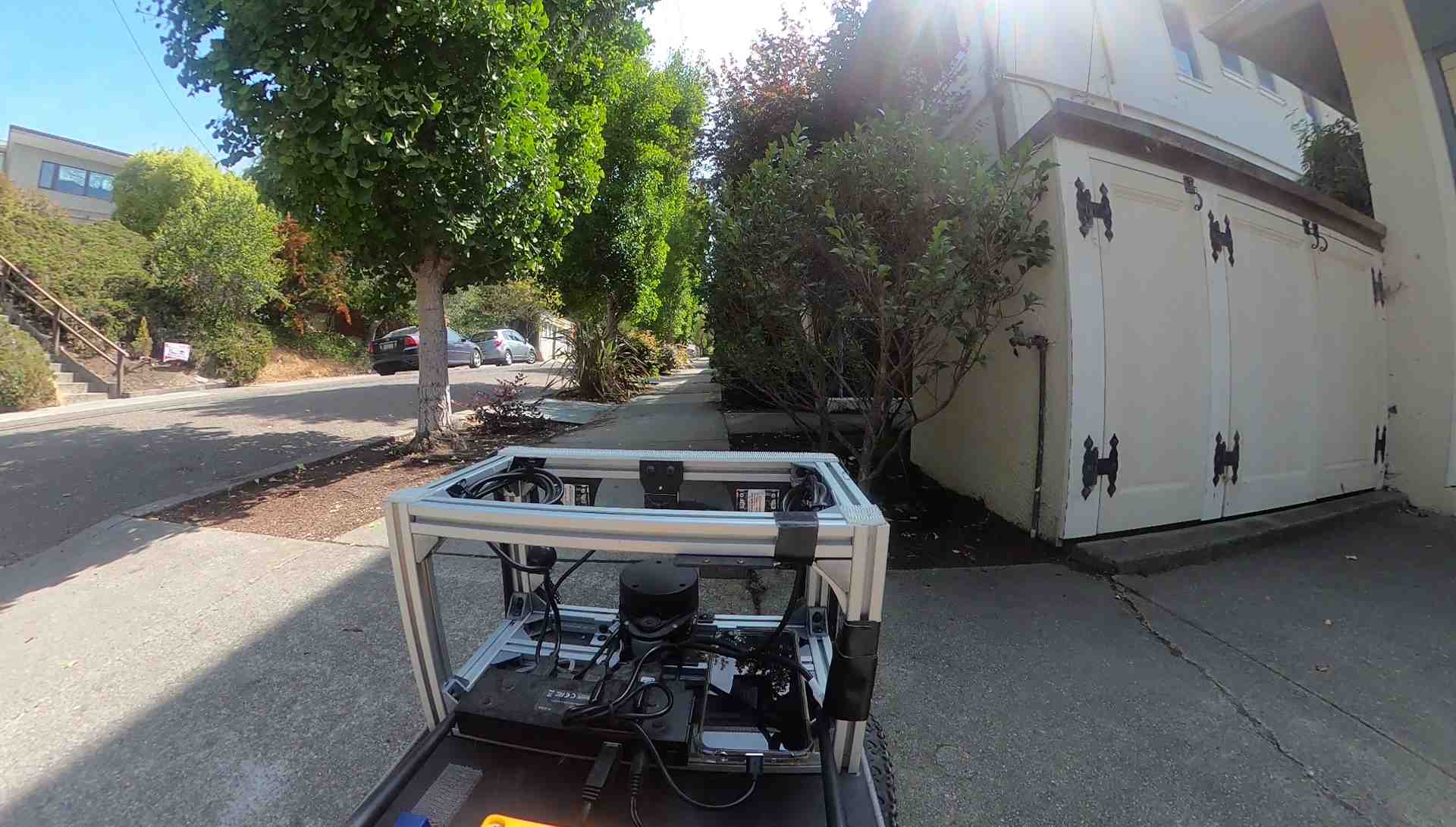} &
	 \includegraphics[width=0.225\textwidth]{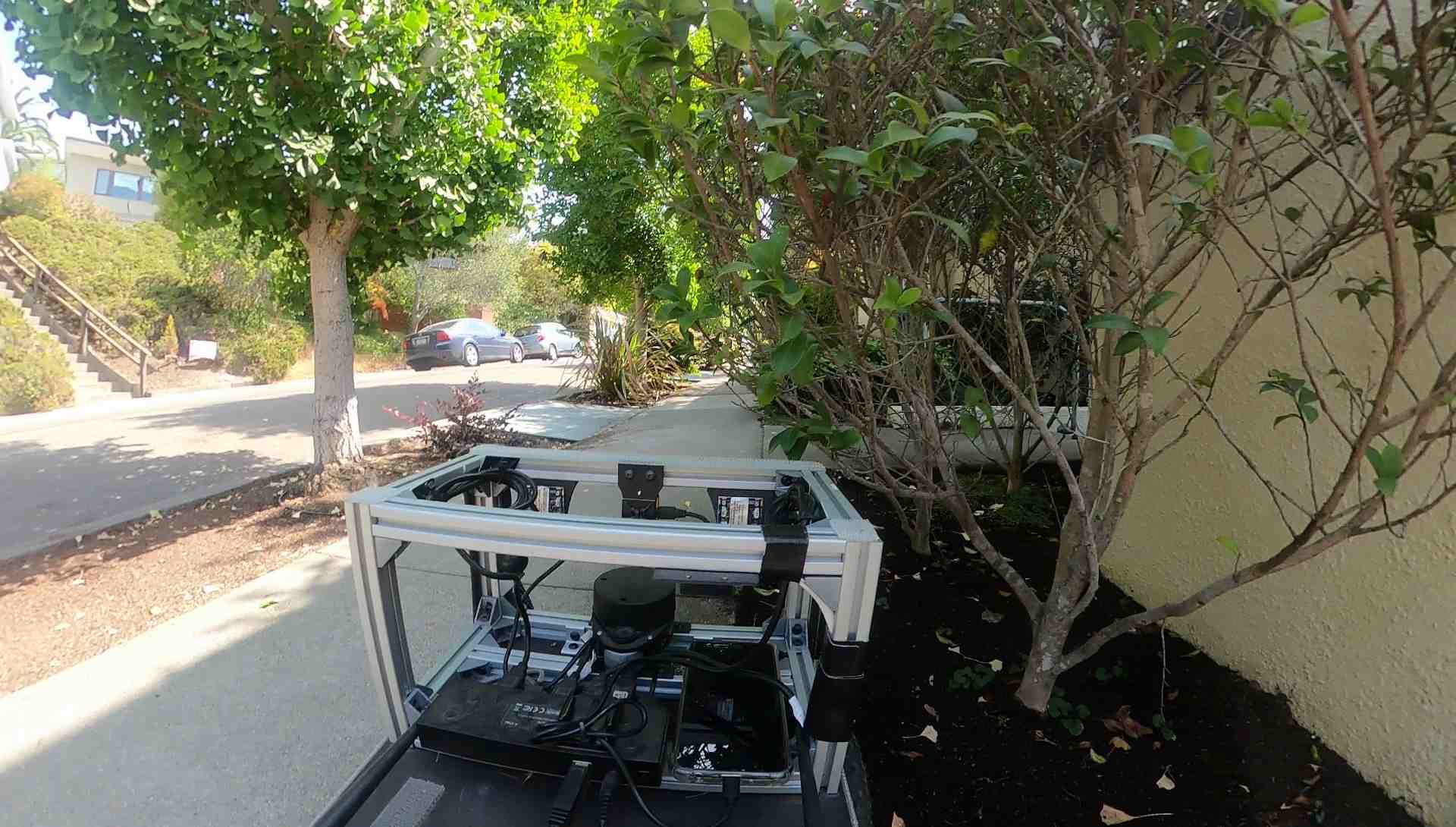} &
	 \includegraphics[width=0.225\textwidth]{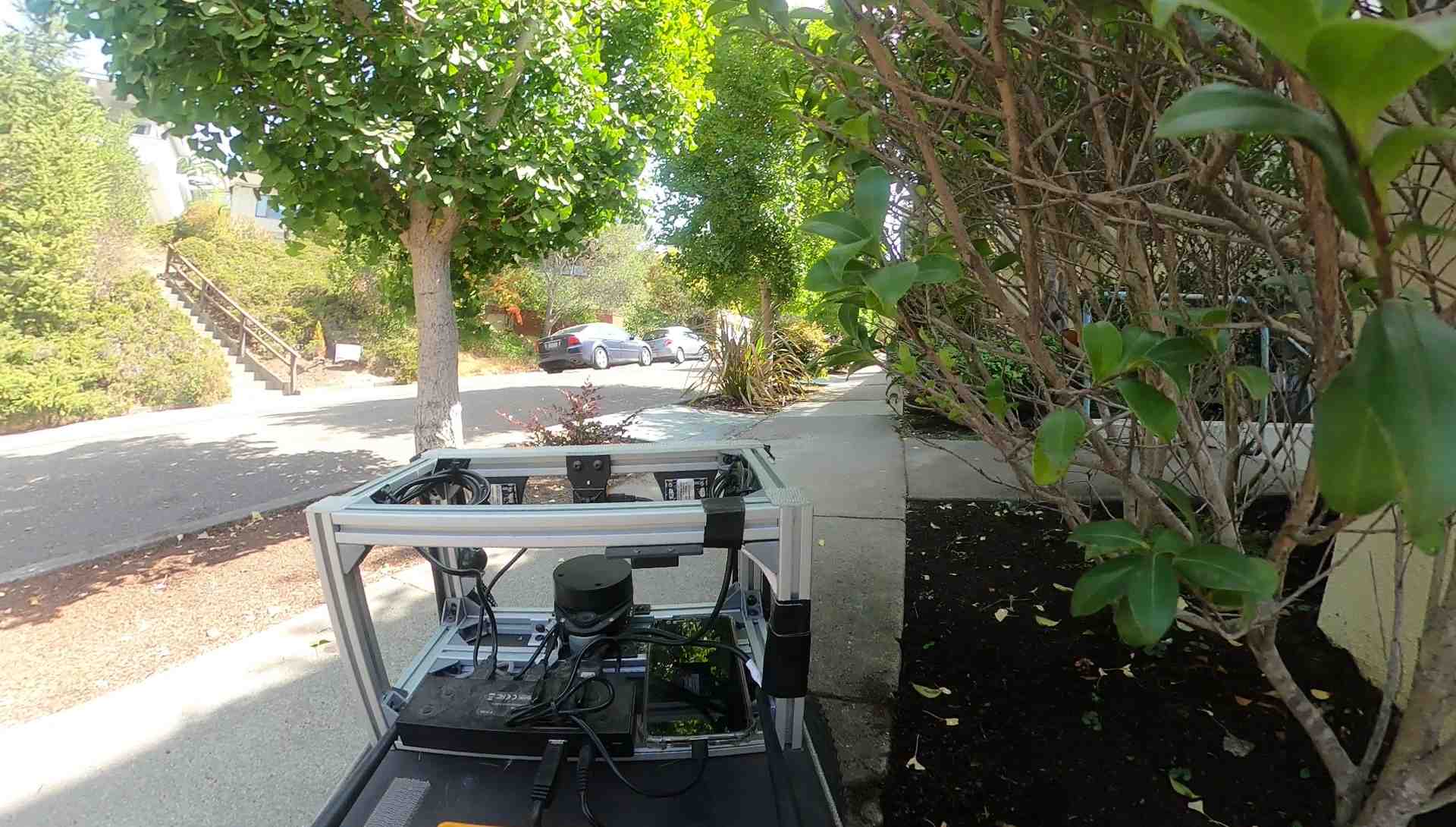} &
	 \includegraphics[width=0.225\textwidth]{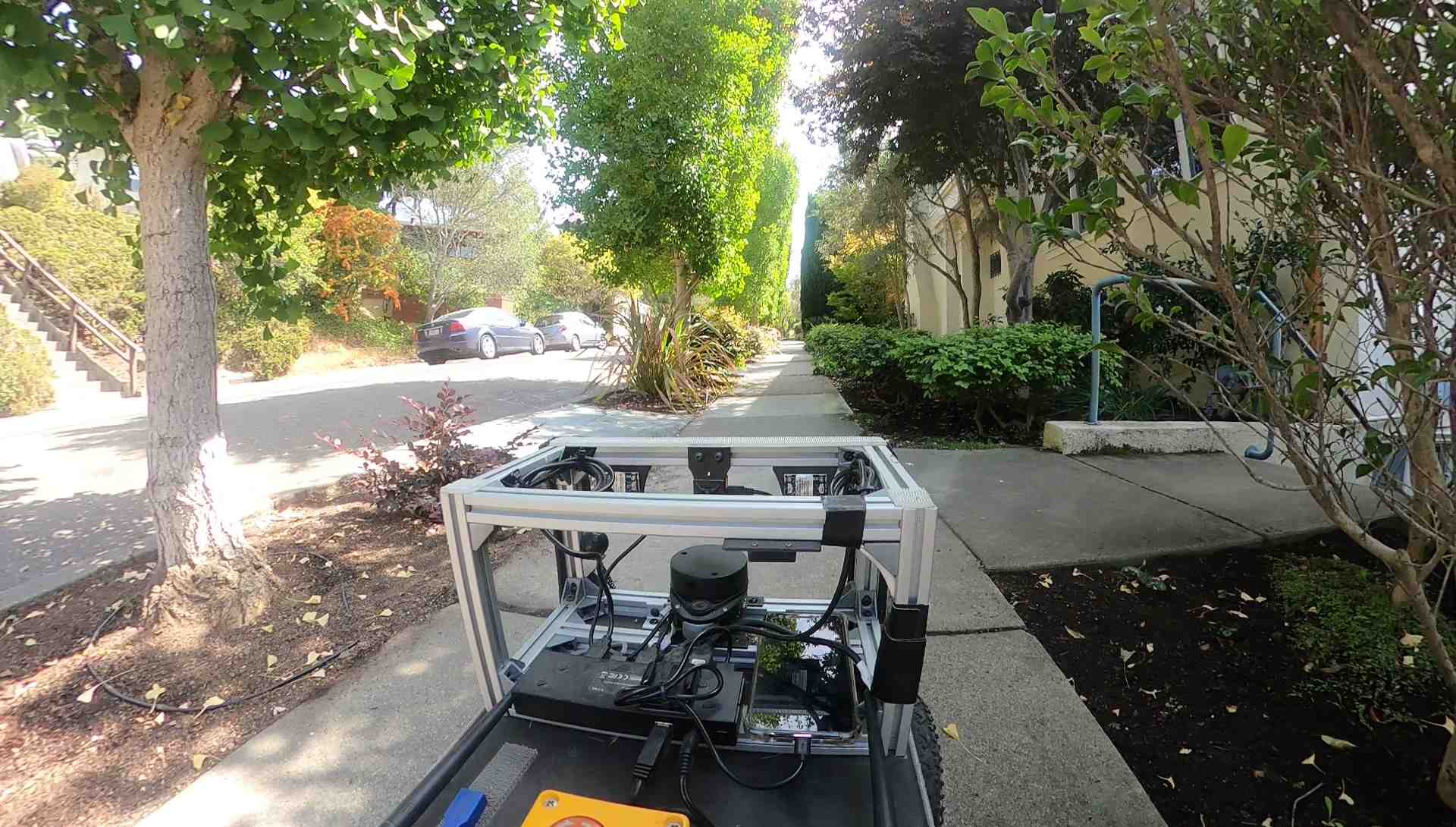} \\
	 
	 	 &
	 \rotatebox[origin=t]{90}{Street\hspace*{-50pt}} &
	 \includegraphics[width=0.225\textwidth]{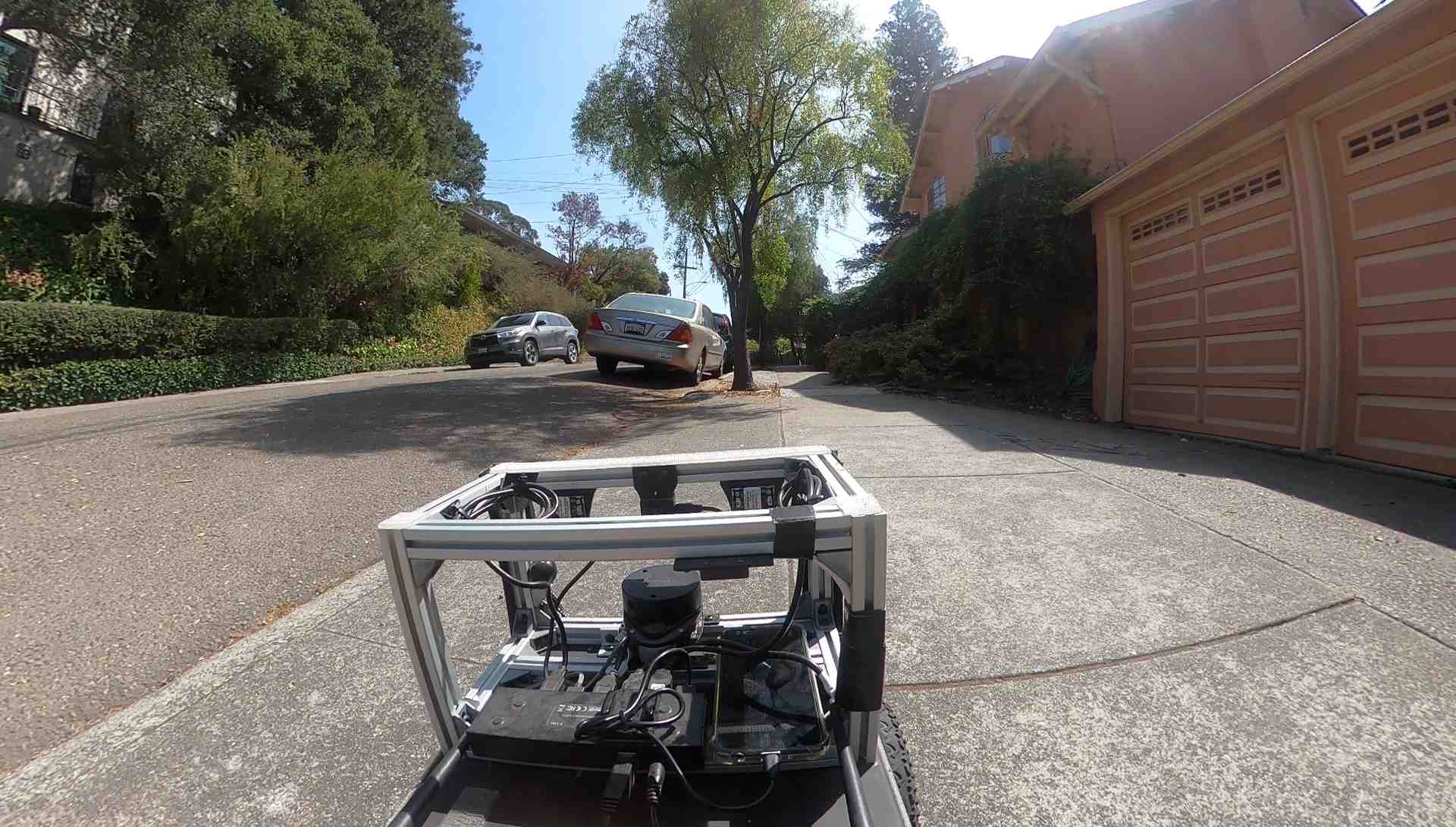} &
	 \includegraphics[width=0.225\textwidth]{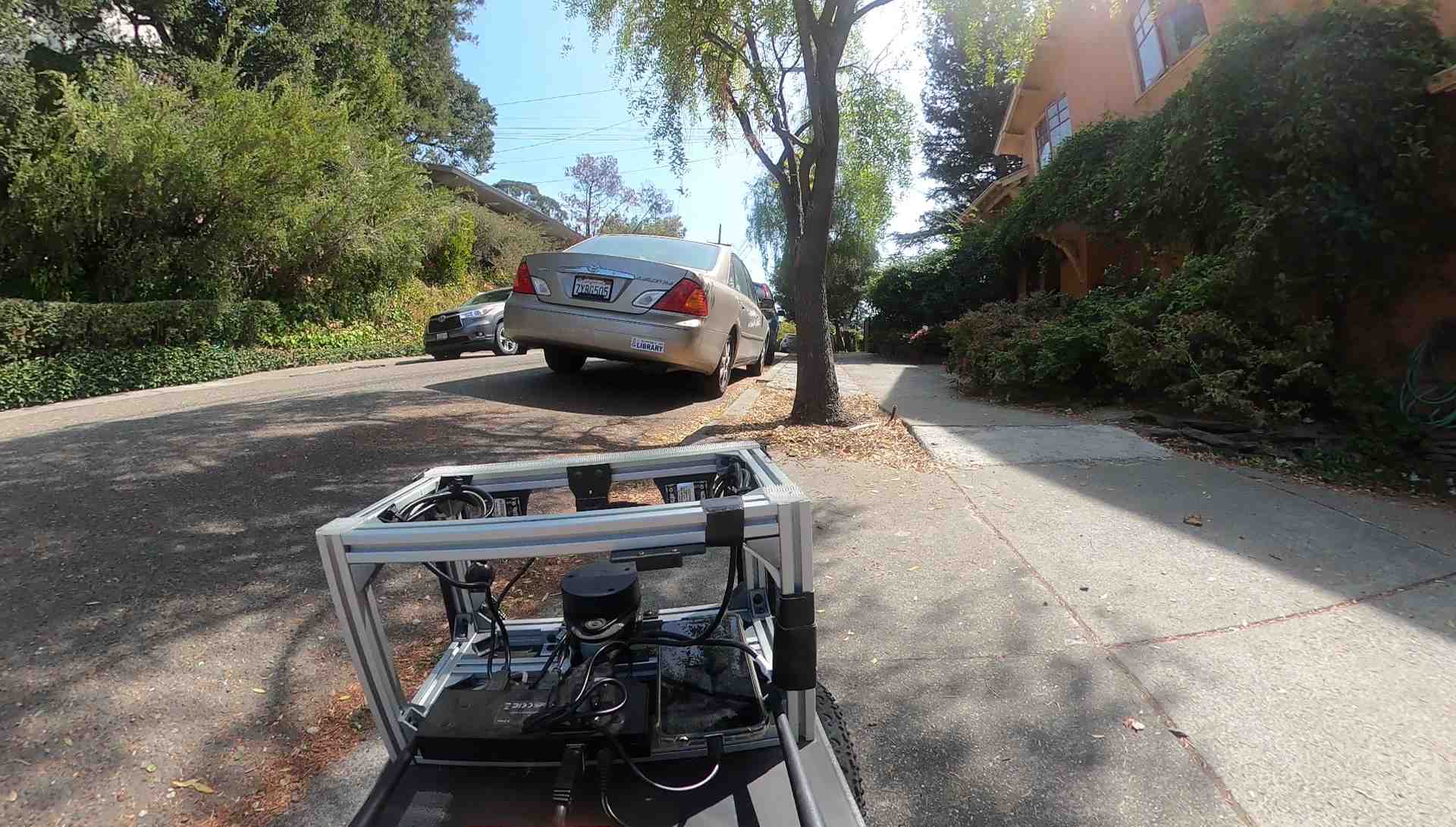} &
	 \includegraphics[width=0.225\textwidth]{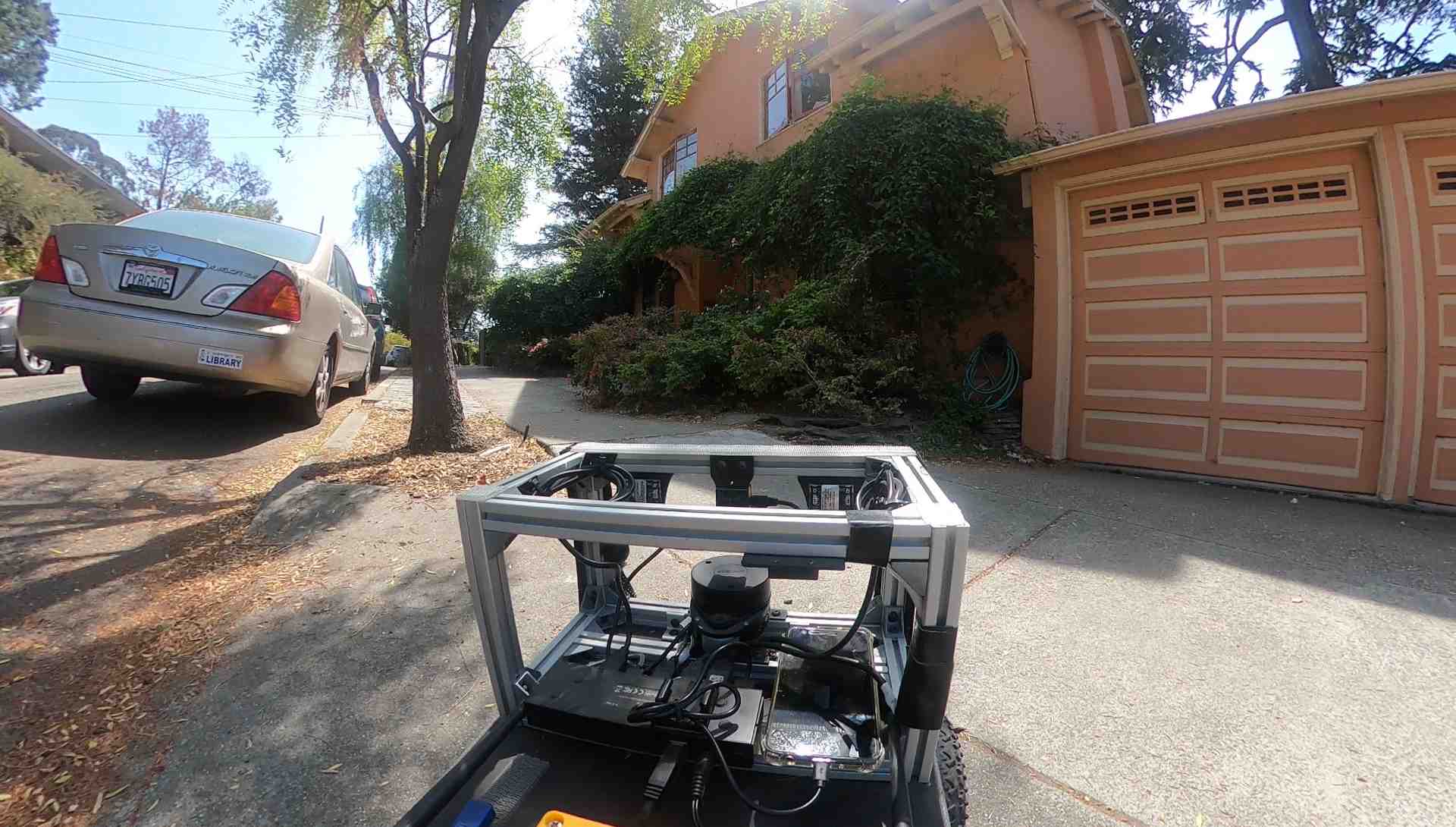} &
	 \includegraphics[width=0.225\textwidth]{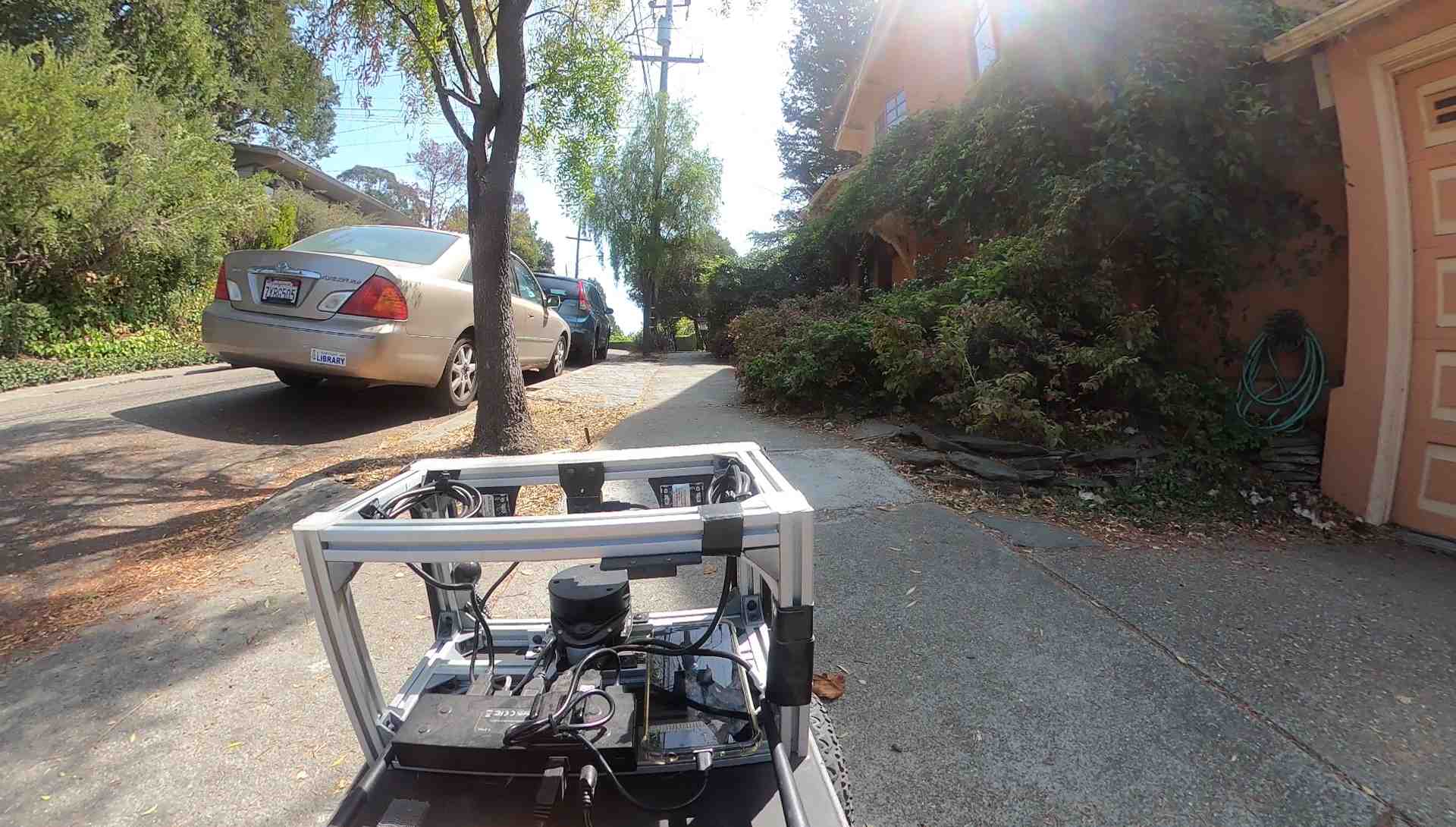} \\
	 
	 	 &
	 \rotatebox[origin=t]{90}{Driveway\hspace*{-50pt}} &
	 \includegraphics[width=0.225\textwidth]{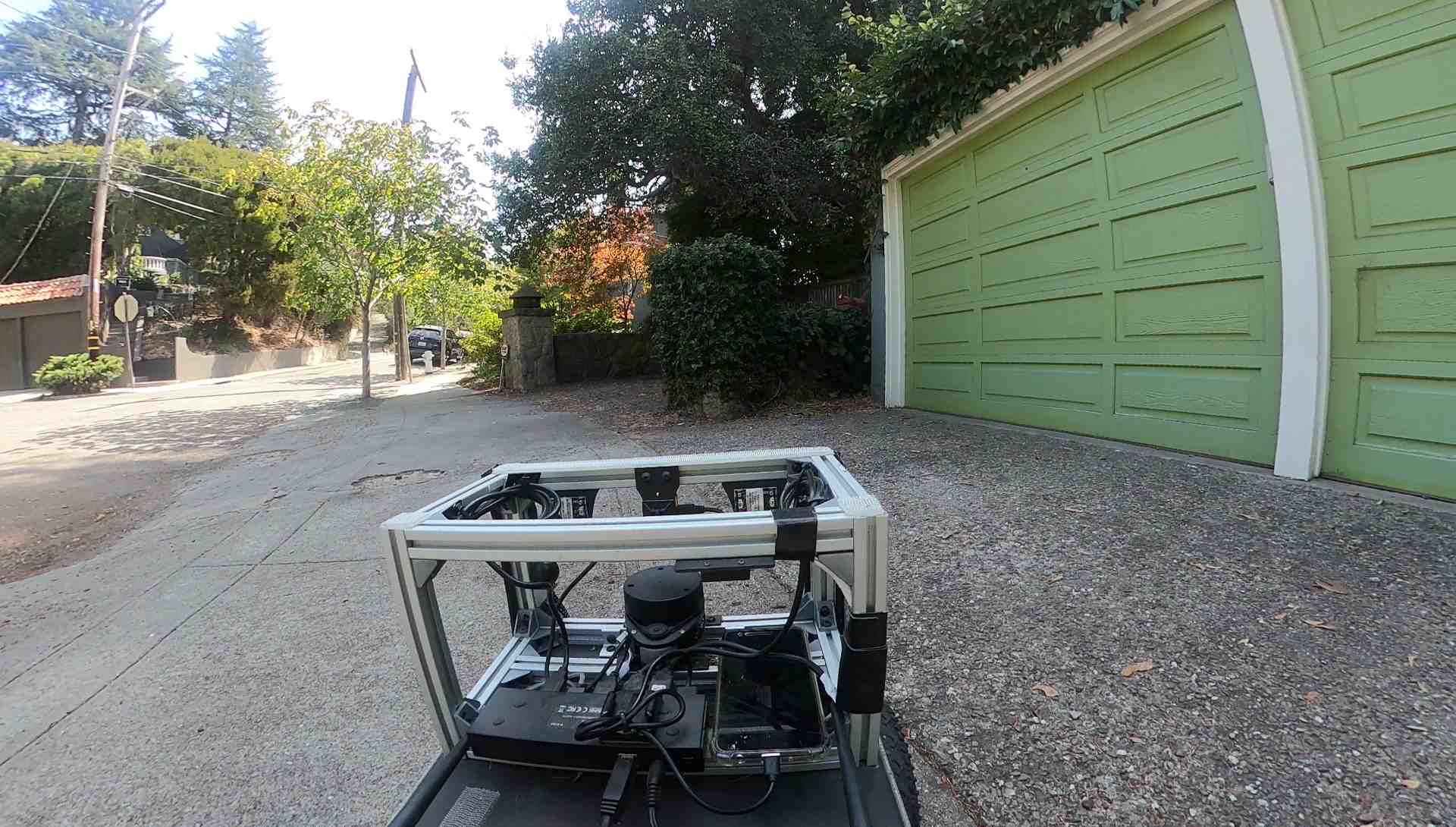} &
	 \includegraphics[width=0.225\textwidth]{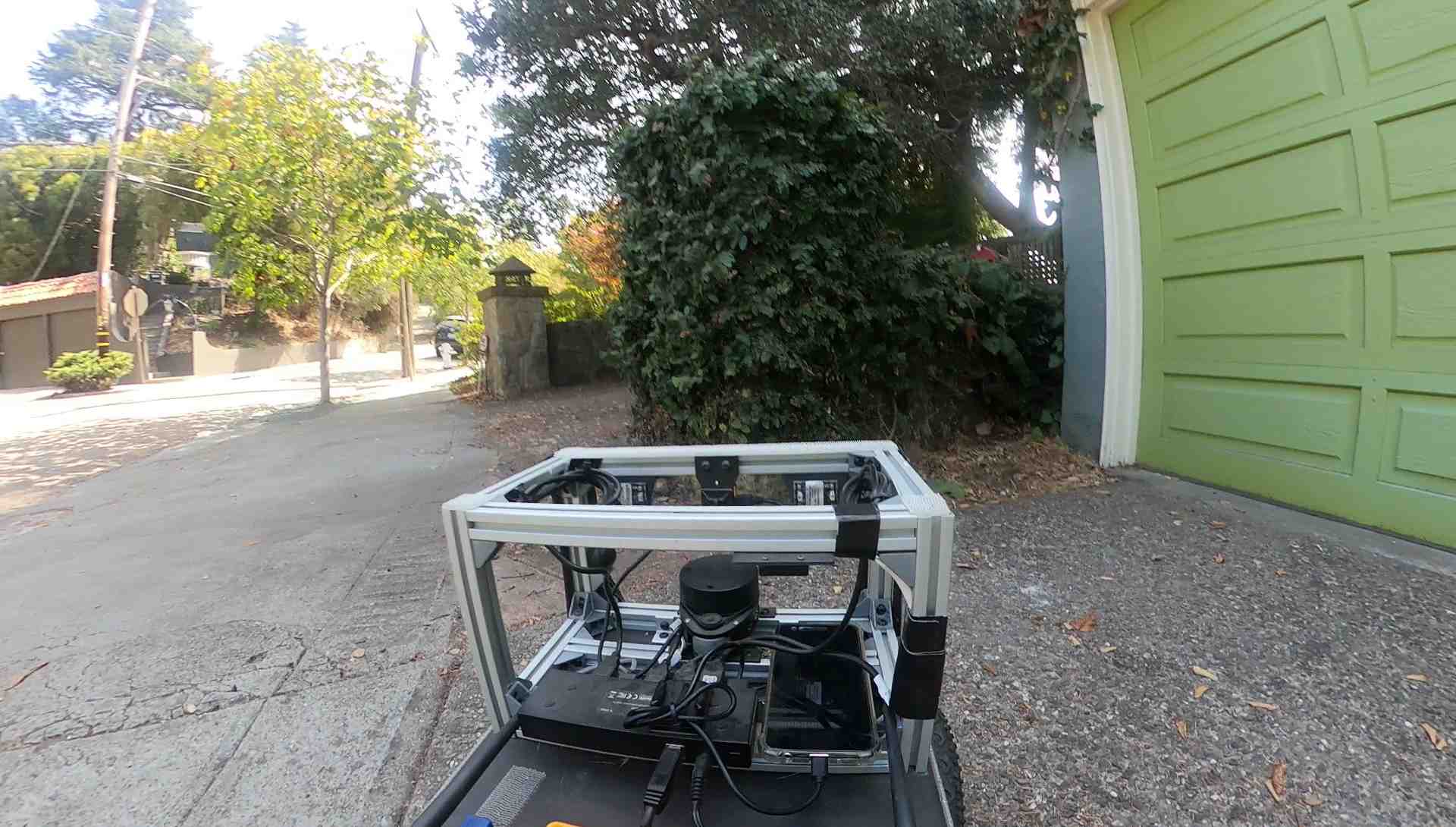} &
	 \includegraphics[width=0.225\textwidth]{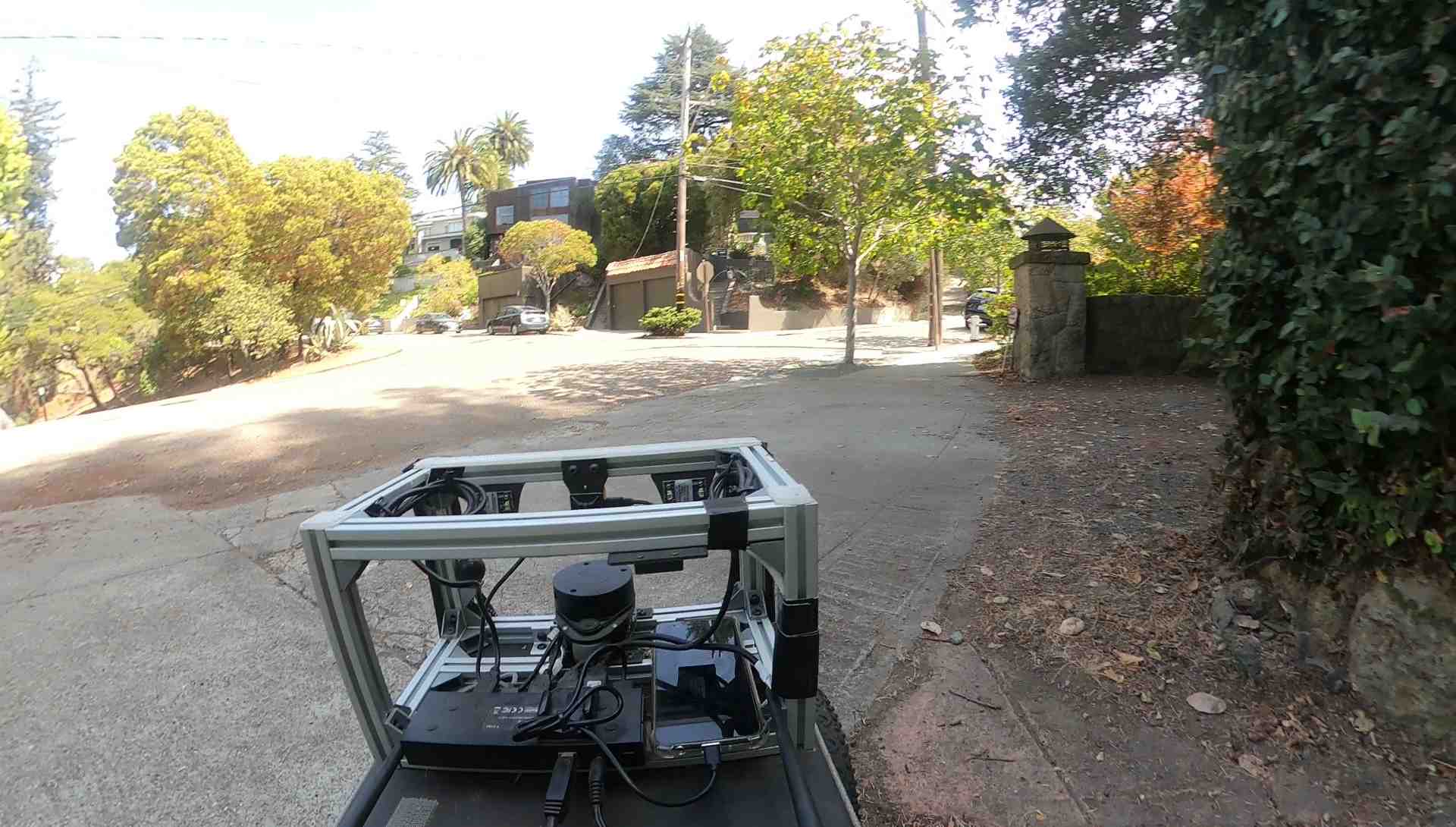} &
	 \includegraphics[width=0.225\textwidth]{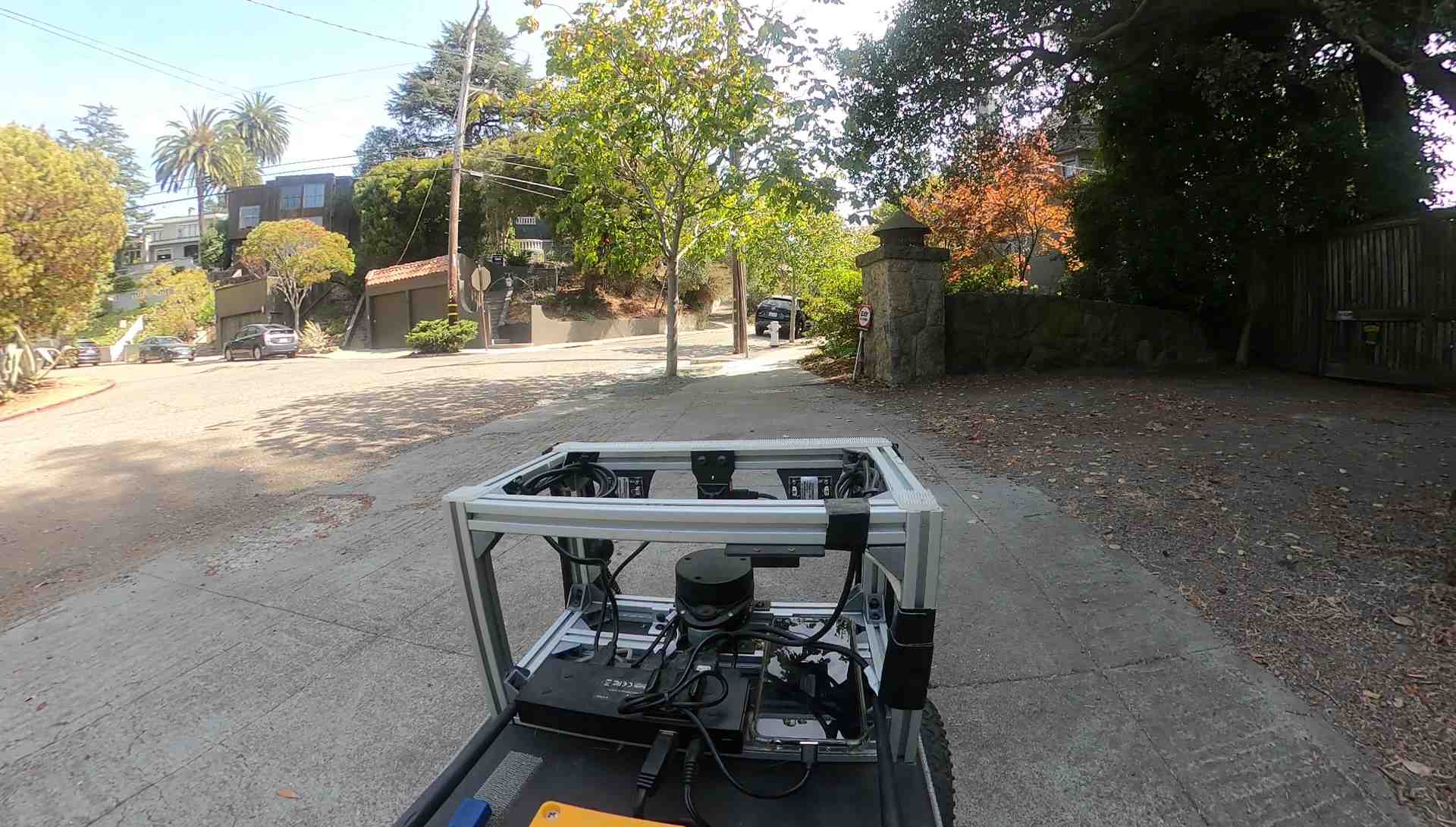} \\
	\end{tabularx}
	\caption{Our \ourmethod approach learns to navigate from disengagements. For the sidewalk navigation task studied in our experiments, there are three types of scenarios that will cause the human overseer to disengage the robot's autonomy mode: colliding with an obstacle, driving into the street, and driving into a driveway. After the robot is disengaged, the human repositions the robot onto the sidewalk and then re-engages autonomy.}
	\label{fig:disengagements}
\end{figure*}
\setlength{\tabcolsep}{\tabledefaultcolspacing}
\renewcommand{\arraystretch}{\tabledefaultrowspacing}
\setcounter{figure}{1}

%%%%%%%%%%%%%%%%%%%%%%%%%%%%%%%%%%%%%%%%%%%%%%%%%%%%%%%%%%%%%%%%%%%%%%%%%%%%%%%%
\section{Related Work}

% robotics
Algorithms for enabling real-world autonomous mobile robot navigation have been extensively studied and applied to a wide range of robotic systems, ranging from ground robots~\cite{Nilsson1968_SRI,Wallace1985_IJCAI,Furgale2010_JFR,Paull2017_ICRA,Starship2020} and aerial drones~\cite{How2008_CSM,Shen2011_ICRA} to full-sized autonomous vehicles~\cite{Montemerlo2008_JFR,Ort2018_ICRA}. These approaches typically follow the paradigm of building a map of the world, localizing the robot within the map, and planning and following paths within the map towards the desired goal~\cite{Fuentes2015_AIR}. While these approaches have enabled recent state-of-the-art results~\cite{Waymo}, these methods still face a number of open challenges, including adverse weather conditions, continuously changing geography, and, perhaps most importantly, failing to learn from prior experience.

% SL
An alternative paradigm for robot navigation are learning-based methods. One common class of these learning-based methods aims to directly learn to predict navigational cues about the environment---such as depth, object detection, and road segmentation---directly from the robot's onboard sensors\cite{Pomerleau1989_NeurIPS,Michels2005_ICML,Fu2018_CVPR,Chang2018_CVPR,Teichmann2018_IV,Wang2019_CVPR}. Although these approaches have been successfully demonstrated, they are exceedingly expensive to train due to the labelling cost, and additional labels do not necessarily lead to improved performance. In contrast, our \ourmethod approach has zero additional cost beyond the already-required human safety driver, and our experiments show \ourmethod directly improves as more data is gathered.

% IL
Another class of learning-based methods is imitation learning, in which a policy is trained to mimic an expert. These approaches have been successfully demonstrated for both ground robots~\cite{Muller2006_NeurIPS,Bojarski2016_arxiv,Pan2018_RSS,Codevilla2018_ICRA} and aerial robots~\cite{Ross2013_ICRA,Giusti2015_RAL,Loquercio2018_RAL}. However, these demonstrations are typically in visually simplistic environments---such as lane following~\cite{Bojarski2016_arxiv,Codevilla2018_ICRA,Loquercio2018_RAL}, hallways~\cite{Loquercio2018_RAL}, and race courses~\cite{Pan2018_RSS}, require injecting noise into the expert's policy~\cite{Ross2013_ICRA,Pan2018_RSS,Codevilla2018_ICRA}---which can be dangerous, and ignore disengagement data~\cite{Muller2006_NeurIPS,Ross2013_ICRA,Giusti2015_RAL,Bojarski2016_arxiv,
Pan2018_RSS}. In contrast, our approach is able to navigate on visually diverse sidewalks, does not require injecting noise during data collection, and can directly leverage disengagement data. Additionally, our experiments demonstrate that our \ourmethod method outperforms behavioral cloning, a standard approach for imitation learning.

% RL
In broader terms, learning from disengagements can be viewed as reinforcement learning~\cite{Sutton1998}, in which a robot learns from trial and error. One class of reinforcement learning methods for robot learning is sim-to-real, in which a control policy is learned in simulation and then executed in the real world~\cite{Sadeghi2017_RSS,Muller2018_CoRL,Hirose2019_RAL}. These sim-to-real approaches are complementary to our \ourmethod approach; the simulation policy can be used to initalize the real-world policy, while our method continues to finetune by learning from disengagements. Other reinforcement learning methods, including ours, learn directly from the robot's experiences~\cite{Riedmiller2007_FBIT,Lipton2016_arxiv,Gandhi2017_IROS,Kahn2017_arxiv,
Richter2017_RSS,Saunders2018_AAMAS,Kendall2019_ICRA,Kahn2020_arxiv}. However, these methods typically assume catastrophic failures are acceptable~\cite{Gandhi2017_IROS,Kahn2020_arxiv} or access to a safe controller~\cite{Kahn2017_arxiv,Richter2017_RSS}, the robot gathers data in a single area over multiple traversals~\cite{Riedmiller2007_FBIT,Gandhi2017_IROS,Kendall2019_ICRA}, on-policy data collection~\cite{Kendall2019_ICRA}, access to a reward signal beyond disengagement~\cite{Riedmiller2007_FBIT}, perform their evaluations in the training environment~\cite{Riedmiller2007_FBIT,Kendall2019_ICRA}, or are only demonstrated in simulation~\cite{Lipton2016_arxiv,Saunders2018_AAMAS}. In contrast, our \ourmethod method is safe because it leverages the existing human-safety driver, learns from off-policy data, does not require retraversing an area multiple times, learns directly from whether the robot is engaged or disengaged, and evaluate in novel, never-before-seen real-world environments. Additionally, we show in our experiments that \ourmethod outperforms~\cite{Kendall2019_ICRA}, a state-of-the-art real world reinforcement learning method for autonomous driving.

%%%%%%%%%%%%%%%%%%%%%%%%%%%%%%%%%%%%%%%%%%%%%%%%%%%%%%%%%%%%%%%%%%%%%%%%%%%%%%%%
\section{Learning to Navigate from Disengagements}

% what's our goal
Our goal is to develop an algorithm, which we call \ourmethod, that enables a mobile robot to autonomously navigate in diverse, real-world environments. An overview of \ourmethod is shown in Fig.~\ref{fig:method-flow}. \ourmethod leverages datasets that are naturally collected while testing autonomous mobile robot systems: the robot's sensor observations---such as camera images, commanded actions---such as the steering angle, and whether the robot autonomy mode was engaged or disengaged. Using this dataset, we then train a convolutional recurrent neural network that takes as input the current observation and a sequence of future commanded actions, and predicts the probability that the robot will be engaged or disengaged in the future. At test time, we can then use this model to plan and execute actions that minimize the probability of disengagement while navigating towards a desired goal location.

% advantages
\ourmethod has several desirable properties. First, our method does not require any additional data beyond what is already collected from testing the autonomous system with a human safety overseer. Second, \ourmethod learns directly from the disengagement data, as opposed to prior methods that use disengagements as an indirect debugging tool. Third, we make minimal assumptions about the robot---just access to the robot's onboard sensors, commanded actions, and whether the autonomy mode was engaged or disengaged; we do not require access to nontrivial information, such as high-definition maps. And lastly, our method continuously improves as the robot is tested, which we demonstrate in our experiments.

% what's to come
In the following sections, we will describe the data collection process, model training, and planning and control, and conclude with a summarizing overview of \ourmethod.

\setcounter{figure}{1}
\begin{figure}[t]
    \centering
    \includegraphics[width=\columnwidth]{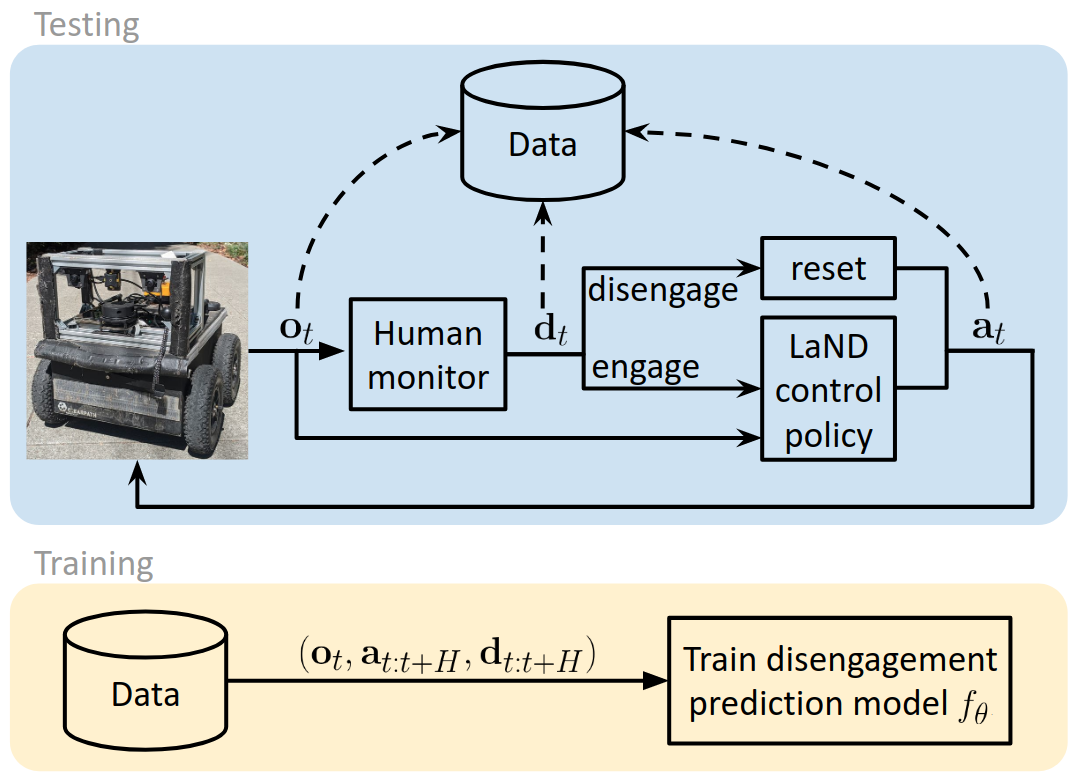}
    \caption{System diagram of our \ourmethod algorithm. In the testing phase, the robot produces sensory observations, such as camera images. Based on these observations, the human monitor determines whether to disengage or engage the robot's autonomy system. If the robot is engaged, our \ourmethod control policy determines which action to execute using the current observation; if the robot is disengaged, the human determines which actions the robot should execute in order to reset. This action is then executed by the robot. While testing, the observations, actions, and disengagements are added to a dataset. In the training phase, this dataset is used to train the disengagement prediction model at the core of the \ourmethod control policy. \ourmethod alternates between training and testing until the control policy reaches satisfactory performance.}
    \label{fig:method-flow}
    \vspace*{-15pt}
\end{figure}

\setcounter{figure}{5}
\begin{figure*}[b]
    \vspace*{-10pt}
    \centering
    \includegraphics[width=\textwidth]{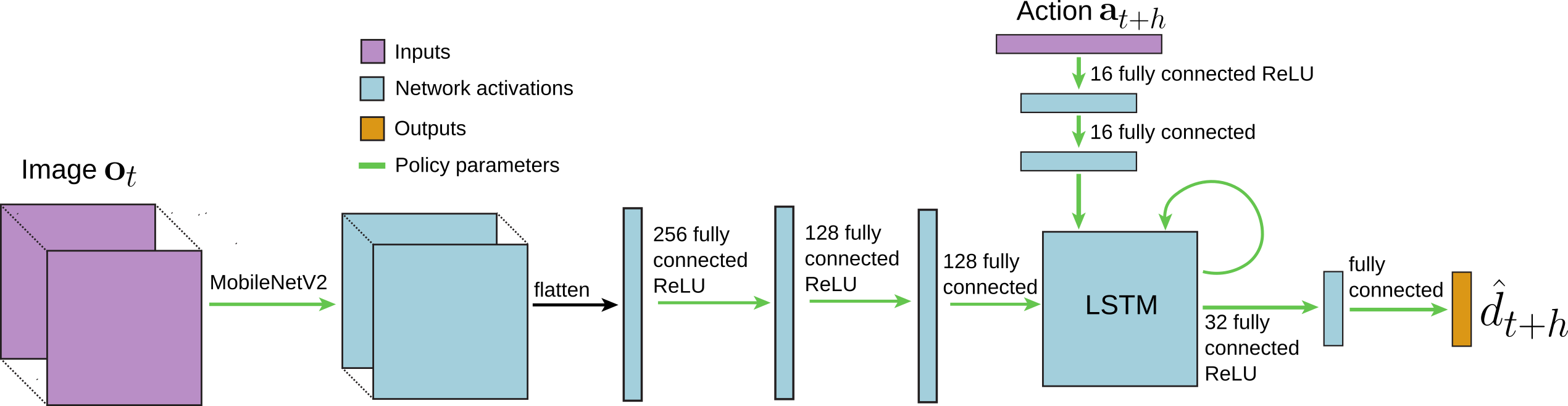}
    \caption{Illustration of the image-based, action-conditioned convolutional recurrent deep neural network at the core of \ourmethod. The network first processes the input image observation using the MobileNetV2~\cite{Sandler2018_CVPR} convolutional neural network, followed by a series of fully connected layers. The output of these image layers serves as the initial hidden state for an LSTM recurrent neural network~\cite{Hochreiter1997_neural}, which sequentially processes each of the $H$ future actions $\ba_{t+h}$ and outputs the corresponding predicted probability of disengagement $\hat{\bd}_{t+h}$. When this model is deployed, these predicted disengagement probabilities enable a planner to plan and execute actions that avoid disengagements.}
    \label{fig:nn}
\end{figure*}

%%%%%%%%%%%%%%%%%%%%%%%%%%%%%%%%%%%%%%
\subsection{Data Collection}

% robot
We start by describing the robot platform used to both collect data and for autonomous navigation. The robot is defined by observation $\bo_t$ gathered by its onboard sensors, action $\ba_t$ which commands the robot, and a binary signal $\bd_t$ indicating if the autonomy mode is disengaged. In our experiments, we use a Clearpath Jackal robot, as shown in Fig.~\ref{fig:method-jackal}. The observation $\bo$ is a $96 \times 192$ RGB image from a front-facing $170^\circ$ field-of-view monocular camera, the action $\ba$
\setcounter{figure}{2}
\begin{wrapfigure}{r}{0.35\columnwidth}
    \centering
    % left bottom right up
    \includegraphics[width=0.35\columnwidth,trim={5cm 0 5cm 0}, clip]{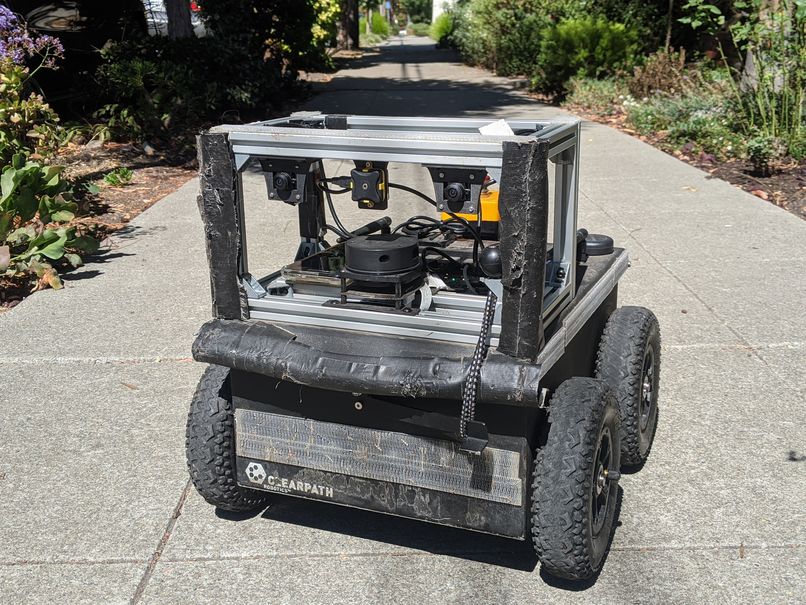}
    \caption{The mobile robot.}
    \label{fig:method-jackal}
    \vspace*{-10pt}
\end{wrapfigure}%
is the desired heading change, and the disengagement signal $\bd$ is conveyed by a human following nearby with a remote control.

% collecting data
Data collection proceeds by having the robot execute an autonomous control policy, such as the \ourmethod planning-based controller described in Sec.~\ref{sec:planning}. A person monitors the robot, and if the robot is in a failure mode or approaching a failure mode, the person disengages autonomy mode. The person then repositions the robot back into a valid state and then re-engages autonomy mode.

% collecting data -- sidewalk
\setcounter{figure}{4}
\begin{wrapfigure}{r}{0.2\columnwidth}
    \vspace*{0pt}
    \centering
    % left bottom right up
    \includegraphics[width=0.2\columnwidth,trim={12cm 0 12cm 0}, clip]{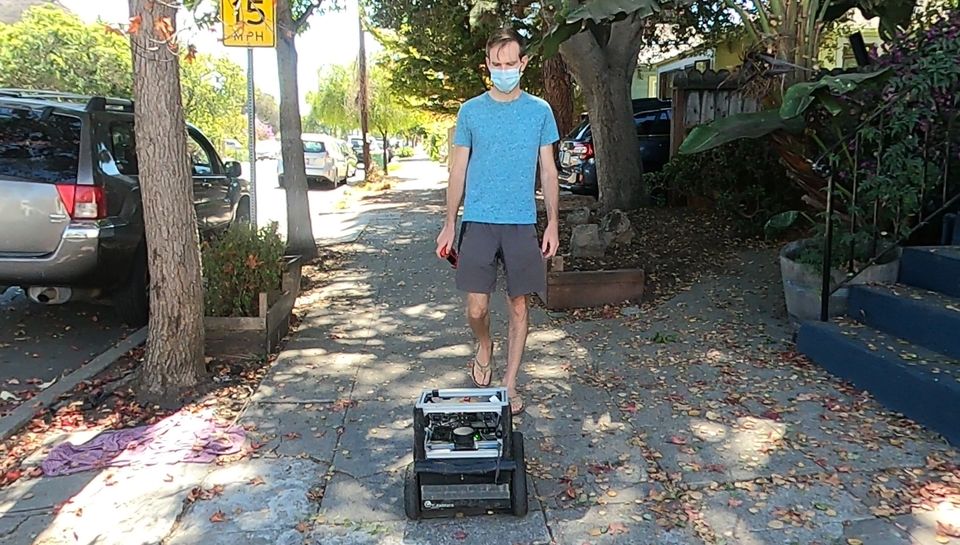}
    \caption{Person monitoring data collection.}
    \label{fig:method-walking-tpv}
    \vspace*{-10pt}
\end{wrapfigure}%
An example of the data collection process for our sidewalk experiment evaluations is shown in Fig.~\ref{fig:disengagements}. The robot proceeds by autonomously driving on the sidewalk. However, if the robot enters one of the three possible failure modes---colliding with an obstacle, driving into the street, or driving into a house's driveway---the person disengages autonomy mode, uses the remote control to reposition the robot onto the sidewalk, and then re-enables autonomy mode.

% saving data
As the robot collects data, the observations, actions, and disengagements $(\bo_t, \ba_t, \bd_t)$ at each time step $t$ are saved every $\dx$ meters into the dataset $\dataset$.

%%%%%%%%%%%%%%%%%%%%%%%%%%%%%%%%%%%%%%
\subsection{Predictive Model}

% model
The learned predictive model takes as input the current observation and a sequence of future actions, and outputs a sequence of future disengagement probabilities. Similarly to~\cite{Kahn2020_arxiv}, we define this model as $f_\theta(\bo_t, \ba_{t:t+H}) \rightarrow \hat{\bd}_{t:t+H}$, which is a function $f$ parameterized by vector $\theta$ that takes as input the current observation $\bo_t$ and a sequence of $H$ future actions $\ba_{t:t+H} = (\ba_t, \ba_{t+1}, ..., \ba_{t+H-1})$, and outputs a sequence of $H$ future predicted disengagement probabilities $\hat{\bd}_{t:t+H} = (\hat{\bd}_t, \hat{\bd}_{t+1}, ..., \hat{\bd}_{t+H-1})$.

% neural network
We instantiate the model as an image-based, action-conditioned convolutional recurrent neural network, as shown in Fig.~\ref{fig:nn}. The network first processes the input image observation using the MobileNetV2~\cite{Sandler2018_CVPR} convolutional neural network, followed by a series of fully connected layers. The output of these image layers serves as the initial hidden state for an LSTM recurrent neural network~\cite{Hochreiter1997_neural}, which sequentially processes each of the $H$ future actions $\ba_{t+h}$ and outputs the corresponding predicted probability of disengagement $\hat{\bd}_{t+h}$.

% loss
The model is trained using the collected dataset to minimize the cross entropy loss between the predicted and ground truth disengagement probabilities
\begin{align}
\loss(\theta, \dataset) = \sum_{(\bo_t, \ba_{t:t+H}, \bd_{t:t+H})} \sum_{h=0}^{H-1} &\lossCE(\hat{\bd}_{t+h}, \bd_{t+h}) \;\; : \nonumber \\
&\hat{\bd}_{t:t+H} = f_\theta(\bo_t, \ba_{t:t+H}), \label{eqn:training-loss}
\end{align}
in which $\lossCE$ is the cross-entropy loss.

% training
The neural network parameters $\theta$ are trained by performing minibatch gradient descent on Eqn.~\ref{eqn:training-loss}. However, we modify the standard minibatch training procedure in two important ways. First, we ensure that half the minibatch contain sequences ending in a disengagement and the other half contain sequences with no disengagements. This rebalancing ensures that disengagement data---which is often a small proportion of the total dataset---is seen often during training. Second, if the sampled time step $t$ is less than $H$ steps from a disengagement, we still sample a full sequence of actions $\ba_{t:t+H}$ and disengagements $\bd_{t:t+H}$ of length $H$ by padding the sequence with (a) actions randomly sampled from the dataset and (b) additional disengagement signals. This artificial disengagement extension scheme assumes that once the robot is disengaged, any action it takes will cause the robot to remain in a disengaged state. This extension scheme is important because it ensures that (a) the model is trained on observations close to disengagements and (b) the training procedure matches the planning procedure---which always plans over a sequence of $H$ actions.

%%%%%%%%%%%%%%%%%%%%%%%%%%%%%%%%%%%%%%
\subsection{Planning and Control}
\label{sec:planning}

Using the trained neural network disengagement prediction model, the robot can plan and execute actions at test time that avoid disengagements while navigating towards a desired goal location.
% cost function
We encode this planning objective with the following cost function
\begin{align}
    \cost(\hat{\bd}_{t:t+H}, \ba_{t:t+H}) = \sum_{h=0}^{H-1} \hat{\bd}_{t+h} + \alpha \cdot \| \ba_{t+h} - \bg \|_2^2. \label{eqn:method-planning-cost}
\end{align}
The first term encourages the robot to avoid disengagements, while the second term encourages the robot to navigate towards a desired goal location; the scalar $\alpha$ is a user-defined weighting between these two terms. The goal location is conveyed through the desired heading vector $\bg$; for example, in our experiments the action is the steering angle, and the goal heading vector is either to turn left, right, or continue straight when the robot encounters a junction. Note that this goal heading does not tell the robot how to navigate, and only provides high-level guidance at junctures; this level of supervision is similar to smartphone driving directions. In our experiments, because there were no junctures, we set $\alpha = 0$, but we maintain this general formulation to show our approach is goal-conditioned.

% planning
Using this cost function, the robot solves solves the following planning problem at each time step
\begin{align}
    \ba^*_{t:t+H} = \arg\min_{\ba_{t:t+H}} &\cost(\hat{\bd}_{t:t+H}, \ba_{t:t+H}) \;\; : \;\; \nonumber \\
    &\hat{\bd}_{t:t+H} = f_\theta(\bo_t, \ba_{t:t+H}), \label{eqn:method-planning-opt}
\end{align}
executes the first action, and continues to plan and execute following the framework of model predictive control~\cite{Camacho2013_springer}.

We solve Eqn.~\ref{eqn:method-planning-opt} using the zeroth order stochastic optimizer from~\cite{Nagabandi2019_CoRL}. This optimizer is more powerful than other zeroth order stochastic optimizers, such as random shooting or the cross-entropy method~\cite{Rubinstein2013_springer}, because it warm-starts the optimization using the solution from the previous time step, uses a soft update rule for the new sampling distribution in order to leverage all of the sampled action sequences, and considers the correlations between time steps. In our experiments, we found this more powerful optimizer was necessary to achieve successful real-time planning and control on our resource-constrained mobile robot. The planner hyperparameters we used were $N=8192, \sigma=1, \beta=0.5, \gamma=50$ as detailed in~\cite{Kahn2020_arxiv}, and we refer the reader to~\cite{Kahn2020_arxiv} and~\cite{Nagabandi2019_CoRL} for implementation details.

%%%%%%%%%%%%%%%%%%%%%%%%%%%%%%%%%%%%%%
\subsection{Algorithm Summary}

We now provide a brief summary of our \ourmethod algorithm (Alg.~\ref{alg:summary}). \ourmethod alternates between two phases: collecting data and training the predictive model.

In the data collection phase, the robot executes actions according to the planning procedure from Eqn.~\ref{eqn:method-planning-opt}. A person monitors the robot, and disengages the robot if it enters a failure mode; if the person does disengage the robot, they then reposition the robot and subsequently re-engage autonomous execution. While collecting data, the current observation, action, and disengagement are saved into the training dataset. In the training phase, the collected dataset is used to train the predicted model by minimizing Eqn.~\ref{eqn:method-planning-opt}.

Although Alg.~\ref{alg:summary} uses our \ourmethod control policy to collect data, we note that any control policy can be used to gather data; in fact, in our experiments we used both \ourmethod and the imitation learning and reinforcement learning comparison methods to gather data. However, we note that the ideal policy for data collection is the \ourmethod policy because this ensures that the collected disengagements are from the failure modes of the \ourmethod policy.

%%%%%%%%%%%%%%%%%%%%%%%%%%%%%%%%%%%%%%%%%%%%%%%%%%%%%%%%%%%%%%%%%%%%%%%%%%%%%%%%
\section{Experiments}

In our experimental evaluation, we study how \ourmethod can learn to navigate from disengagement data in the context of a sidewalk navigation task, and compare our approach to state-of-the-art imitation learning and reinforcement learning approaches. Videos, code, and other supplemental material are available on our website~\footnote{\url{https://sites.google.com/view/sidewalk-learning}}

Our dataset consists of 17.4 km of sidewalks gathered over 6 hours. Data was saved every $\dx=0.5$ meters, and therefore the dataset has 34,800 data points, and contains 1,926 disengagements. Although the amount of data gathered may seem significant, (1) this data is already being collected while testing the robot, (2) the robot requires less human disengagements as it gathers more data and trains, and (3) this dataset is significantly smaller than those typically used in computer vision~\cite{Deng2009_CVPR} and reinforcement learning~\cite{Hessel2018_AAAI} algorithms.

% videos (thumbnails) of our method working and IL failing
\setcounter{figure}{8}
\setlength{\tabcolsep}{2.5pt}
\renewcommand{\arraystretch}{1}
\begin{figure*}[b]
    \vspace*{-10pt}
	\begin{tabularx}{\textwidth}{l *{4}{Y} }

	\rotatebox[origin=t]{90}{{\small Behavioral cloning}\hspace*{-75pt}} &
	\includegraphics[width=0.23\textwidth]{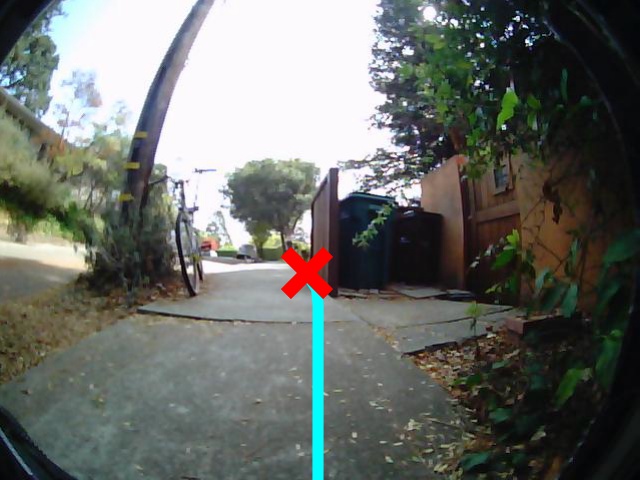} &
	\includegraphics[width=0.23\textwidth]{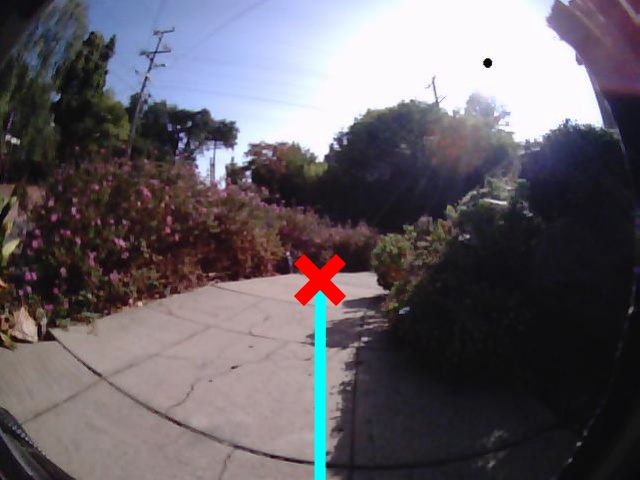} &
	\includegraphics[width=0.23\textwidth]{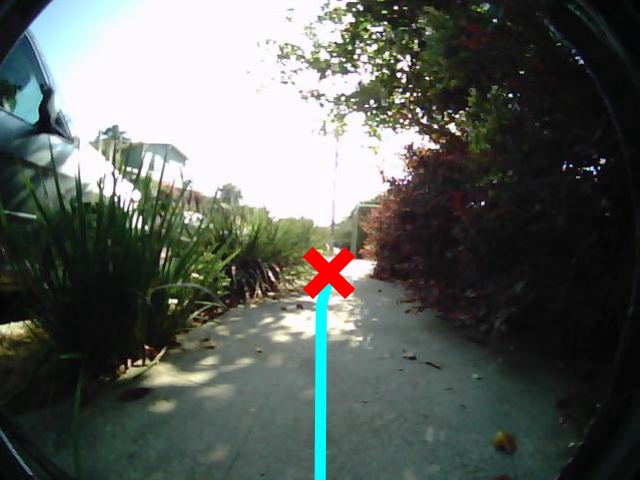} &
	\includegraphics[width=0.23\textwidth]{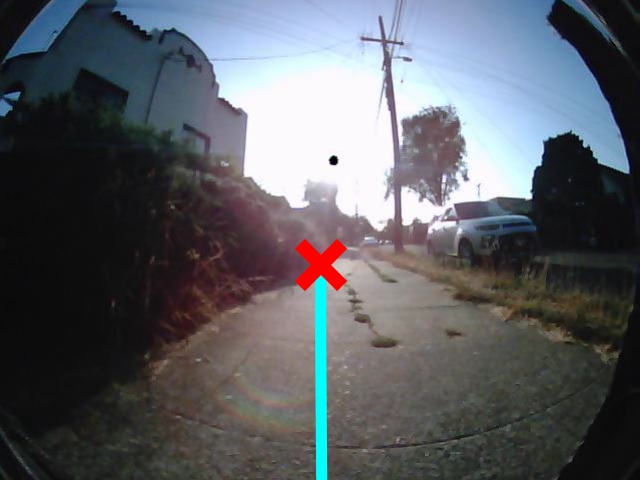} \\

	\rotatebox[origin=t]{90}{\textbf{{\small \ourmethod (ours)}\hspace*{-75pt}}} &
	\includegraphics[width=0.23\textwidth]{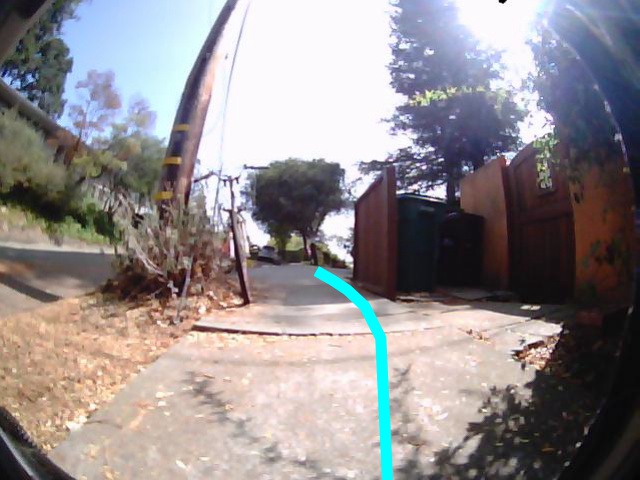} &
	\includegraphics[width=0.23\textwidth]{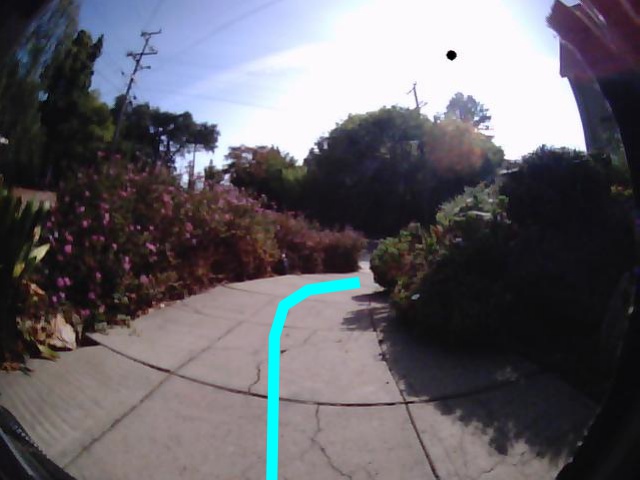} &
	\includegraphics[width=0.23\textwidth]{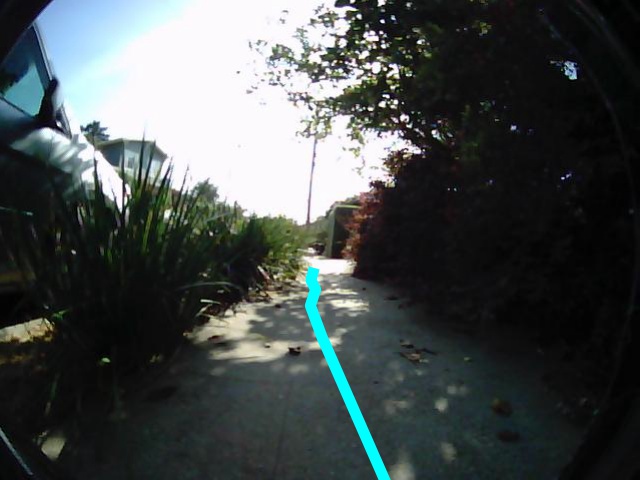} &
	\includegraphics[width=0.23\textwidth]{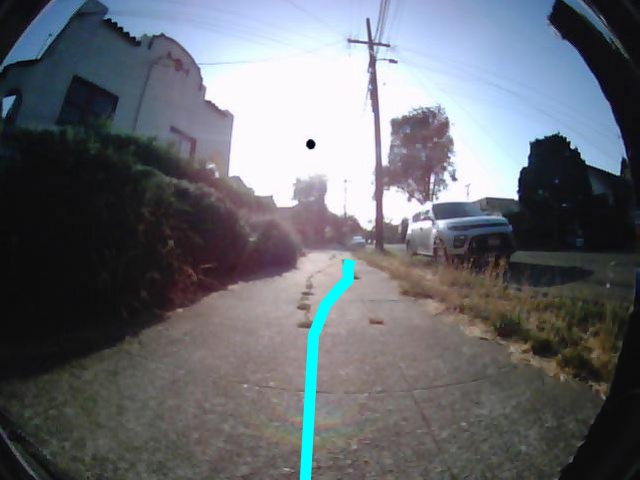}
	
	\end{tabularx}
	\vspace*{-15pt}
	\caption{Qualitative comparison of our \ourmethod method versus the best performing prior approach (behavioral cloning) in scenarios containing parked bicycles, dense foliage, sun glare, and sharp turns. Our approach successfully navigates these scenarios, while imitation learning is unable to and crashes.}
	\label{fig:ours_vs_il_thumbnails}
\end{figure*}
\setlength{\tabcolsep}{\tabledefaultcolspacing}
\renewcommand{\arraystretch}{\tabledefaultrowspacing}

\begin{algorithm}[t]
\caption{Learning to Navigate from Disengagements}
\label{alg:summary}
\begin{algorithmic}[1]
\STATE initialize dataset $\mathcal{D} \leftarrow \emptyset$
\STATE randomly initialize learned parameter $\theta$
\WHILE{not done}
    \WHILE{collecting data}
    	\STATE get current observation $\bo_t$ from sensors
	    \STATE solve Eqn.~\ref{eqn:method-planning-opt} using $f_\theta$ and  $\bo_t$ to get the\\planned action sequence $\ba^*_{t:t+H}$
	    \STATE execute the first action $\ba^*_t$
    	\STATE get current disengagement $\bd_t$
    	\STATE add $(\bo_t, \ba^*_t, \bd_t)$ to $\dataset$
    	\IF{$\bd_t$ is disengaged}
    		\STATE let human execute reset maneuver and re-engage autonomy
    	\ENDIF
    \ENDWHILE
    \STATE use $\mathcal{D}$ to train predictive model $f_\theta$\\by minimizing Eqn.~\ref{eqn:training-loss}
\ENDWHILE
\end{algorithmic}
\end{algorithm}

We evaluated \ourmethod in comparison to two other methods:
\begin{enumerate}
    \item \textit{Behavioral cloning}: a common imitation learning approach used by many state-of-the-art navigation methods~\cite{Bojarski2016_arxiv,Codevilla2018_ICRA,Loquercio2018_RAL}.
    \item \textit{Kendall et. al.~\cite{Kendall2019_ICRA}}: a reinforcement learning algorithm which first learns a compressed representation of the training images using a VAE~\cite{Kingma2014_ICLR}, and learns a control policy from this compressed representation using the DDPG reinforcement learning algorithm~\cite{Lillicrap2016_ICLR}. Kendall et al.~\cite{Kendall2019_ICRA} did not provide source code, so we therefore used existing VAE~\footnote{\url{www.tensorflow.org/tutorials/generative/cvae}} and DDPG~\footnote{\url{www.github.com/rail-berkeley/d4rl_evaluations}} implementations.
\end{enumerate}
All methods, including ours, were trained on the same dataset. For behavioral cloning, data within 2 meters of a disengagement was not used for training. For Kendall et. al.~\cite{Kendall2019_ICRA}, the reward was -1 for a disengagement, and 0 otherwise.

We compare against these methods because, to the best of our knowledge, they are representative of state-of-the-art methods in imitation learning and reinforcement learning for such tasks. We note, however, that we were unable to perfectly replicate these algorithms because they contained assumptions that violated our problem statement; for example, many of the algorithms injected exploration noise into the data collection policy. Nevertheless, we believe our evaluation accurately reflects current work in the context of our realistic problem statement and real world experimental evaluation.

\begin{table}[t]
	\centering
    \begin{tabular}{|l|c|}
         \hline \multicolumn{1}{|c|}{Method} & \specialcell{Avg. distance until\\disengagement (meters)}  \\
         \hline Behavioral cloning (e.g.,~\cite{Bojarski2016_arxiv,Codevilla2018_ICRA,Loquercio2018_RAL}) & 13.4 \\
         \hline Kendall et. al.~\cite{Kendall2019_ICRA} & 2.0 \\
         \hline \textbf{\ourmethod (ours)} & \textbf{87.5} \\
         \hline
    \end{tabular}
    \caption{Experimental evaluation on 2.3 km of never-before-seen sidewalks (Fig.~\ref{fig:teaser}). Our \ourmethod approach is better able to navigate these sidewalks, travelling $6.5 \times$ further before disengagement compared to the next best method.}
    \label{tab:experiments-generalization}
    \vspace*{-15pt}
\end{table}

\setcounter{figure}{6}
\begin{figure}[t]
	\centering
    \includegraphics[width=\columnwidth]{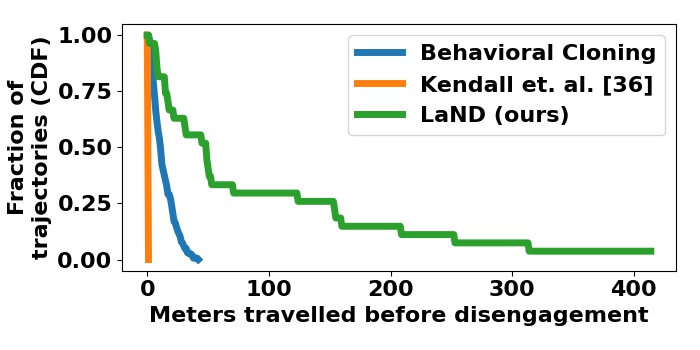}
    \vspace*{-22pt}
    \caption{Experimental evaluation on 2.3 km of never-before-seen sidewalks (Fig.~\ref{fig:teaser}). The plot shows the fraction of trajectories---defined as a continuous episode of engaged autonomy---which travelled a certain distance before a disengagement. Methods closer to the top right are better because this indicates a longer distance travelled before disengagement. Our \ourmethod approach is able to travel farther before disengagement: 33\% of the trajectories travelled further than 50 meters, including a trajectory of over 400 meters. In contrast, none of the prior methods were able to travel more than 50 meters before disengagement.}
    \label{fig:experiments-generalization-cdf}
    \vspace*{-15pt}
\end{figure}

\begin{figure}[t]
    \centering
    \includegraphics[height=0.112\textheight]{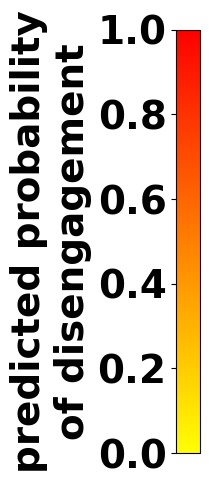}
    \hfill
    \includegraphics[height=0.112\textheight]{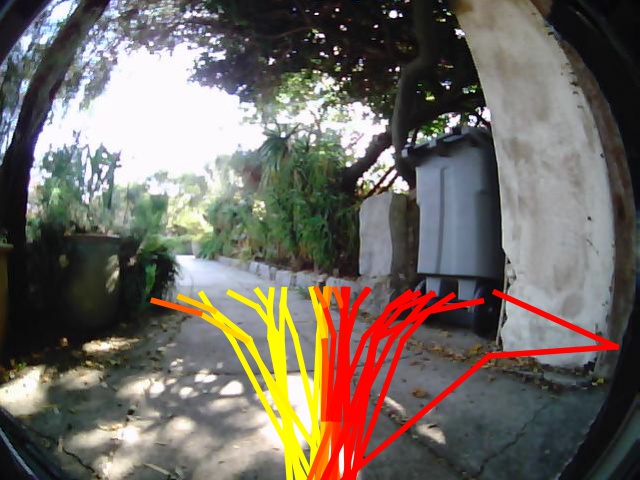}
    \hfill
    \includegraphics[height=0.112\textheight]{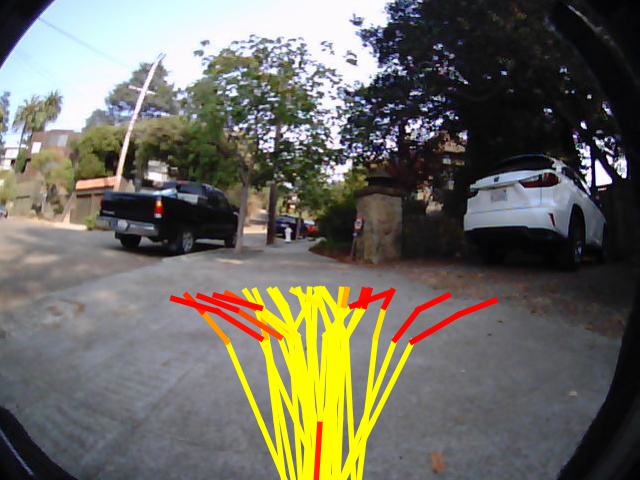}
    \vspace*{-10pt}
    \caption{Visualization of planning with the disengagement predictive model. Each image shows the candidate paths considered during planning, and are color coded according to their predicted probability of disengagement. These visualizations show the learned model can accurately predict which action sequences would lead to disengagements, including driving into obstacles (left) or streets or driveways (right).}
    \label{fig:action_selection}
    \vspace*{-15pt}
\end{figure}

We evaluated all approaches on 2.3 km of sidewalks not present in the training data, as shown in Fig.~\ref{fig:teaser}. Table~\ref{tab:experiments-generalization} shows that \ourmethod is better able to navigate sidewalks compared to the other approaches, traveling $6.5 \times$ further on average before disengagement compared to the next best approach. Fig.~\ref{fig:experiments-generalization-cdf} shows a more detailed per-trajectory analysis. None of the methods besides \ourmethod were able to travel further than 50 meters before disengagement. \ourmethod was able to travel further than 50 meters before disengagement for 33\% of the trajectories, and could sometimes travel up to 400 meters. Fig.~\ref{fig:ours_vs_il_thumbnails} shows example challenge scenarios---including a parked bicycle, dense foliage, sun glare, and sharp turns---in which \ourmethod successfully navigated, while the best performing comparative approach (imitation learning) failed.

\begin{table}[b]
    \vspace*{-15pt}
    \centering
    \begin{tabular}{|l|c|}
         \hline \multicolumn{1}{|c|}{Method} & \specialcell{Avg. distance until\\disengagement (meters)} \\
         \hline \ourmethod & 101.2 \\
         \hline \ourmethod with finetuning & 218.3 \\
         \hline
    \end{tabular}
    \caption{Experimental demonstration of our \ourmethod improving as more data is gathered. We first collected data using our \ourmethod control policy on 1.3 km of never-before-seen sidewalk. We then finetuned our method on this additional data, and evaluated this finetuned model on the same 1.3 km of sidewalk. Our method improves by over $2 \times$ when finetuned on the collected data, showing that our approach is able to continue to improve as more data is gathered.}
    \label{tab:experiments-train-on-test}
\end{table}

To demonstrate that our approach is able to learn which action sequences lead to disengagements and which do not, we visualized the planner in Fig.~\ref{fig:action_selection}. This visualization shows the model has learned that actions which lead to collisions, driving into the street, or driving into a driveway will lead to a disengagement.

We also investigated the ability of \ourmethod to continuously improve as more data is gathered. In this controlled experiment, we first ran our method on 1.3 km of sidewalk. We then trained our method on this additional data---which included mostly engagements, but also a few disengagements---and evaluated this finetuned model on the same 1.3 km of sidewalk. Table~\ref{tab:experiments-train-on-test} shows that our method improves by over $2 \times$ when finetuned on the collected data, showing that our approach is able to continue to improve as more data is collected.

%%%%%%%%%%%%%%%%%%%%%%%%%%%%%%%%%%%%%%%%%%%%%%%%%%%%%%%%%%%%%%%%%%%%%%%%%%%%%%%%
\section{Discussion}

We presented \ourmethod, a method for learning navigation policies from disengagements. \ourmethod directly leverages disengagement data by learning a model that predicts which actions lead to disengagements given the current sensory observation. This predictive disengagement model can then be used for planning and control to avoid disengagements. Our results demonstrate \ourmethod can successfully learn to navigate in diverse, real world sidewalk environments, demonstrating that disengagements are not only useful as a tool for testing, but also directly for learning to navigate.

% One limitation and its solution (a real limitation, not a made-up one)
% the discussion of bootstrapping the method from imitation (e.g., why bootstrapping is a problem, how to solve)
% one "future work" that is really just a way to point out the many cool things that the method enables.
Although \ourmethod can learn to navigate from disengagements, requiring a person to constantly monitor the robot is both tedious and costly. Investigating methods which enable the robot to know when to ask the human to monitor---such as when the robot is uncertain---could minimize the human cost. Also, while \ourmethod can leverage data collected while testing the autonomous robot, there are also many large navigation datasets containing expert data. Investigating how \ourmethod can be combined with imitation learning approaches into a unified framework trained using disengagement and expert data could improve the final navigation policy. Lastly, while we presented \ourmethod as a standalone navigation system, exploring how \ourmethod could be combined with existing navigation planning systems could enable \ourmethod to \textit{patch} the failure modes of these systems. We believe that solving these and other challenges is crucial for enabling mobile robots to successfully navigate in complex, real world environments.

%%%%%%%%%%%%%%%%%%%%%%%%%%%%%%%%%%%%%%%%%%%%%%%%%%%%%%%%%%%%%%%%%%%%%%%%%%%%%%%%
\bibliographystyle{IEEEtran}
\bibliography{references}

\begin{thebibliography}{10}
\providecommand{\url}[1]{#1}
\csname url@rmstyle\endcsname
\providecommand{\newblock}{\relax}
\providecommand{\bibinfo}[2]{#2}
\providecommand\BIBentrySTDinterwordspacing{\spaceskip=0pt\relax}
\providecommand\BIBentryALTinterwordstretchfactor{4}
\providecommand\BIBentryALTinterwordspacing{\spaceskip=\fontdimen2\font plus
\BIBentryALTinterwordstretchfactor\fontdimen3\font minus
  \fontdimen4\font\relax}
\providecommand\BIBforeignlanguage[2]{{%
\expandafter\ifx\csname l@#1\endcsname\relax
\typeout{** WARNING: IEEEtran.bst: No hyphenation pattern has been}%
\typeout{** loaded for the language `#1'. Using the pattern for}%
\typeout{** the default language instead.}%
\else
\language=\csname l@#1\endcsname
\fi
#2}}

\bibitem{DisengagementReport2020}
\BIBentryALTinterwordspacing
T.~Verge. (2020) Everyone hates california’s self-driving car reports.
  [Online]. Available:
  \url{https://www.theverge.com/2020/2/26/21142685/california-dmv-self-driving-car-disengagement-report-data}
\BIBentrySTDinterwordspacing

\bibitem{Nilsson1968_SRI}
N.~Nilsson, B.~Raphael, and S.~Wahlstrom,
  ``\href{http://citeseerx.ist.psu.edu/viewdoc/download?doi=10.1.1.184.8908&rep=rep1&type=pdf}{Application
  of Intelligent Automatat to Reconnaissance},'' 1968.

\bibitem{Wallace1985_IJCAI}
R.~S. Wallace, A.~Stentz, C.~E. Thorpe, H.~P. Moravec, W.~Whittaker, and
  T.~Kanade,
  ``\href{https://www.ijcai.org/Proceedings/85-2/Papers/086.pdf}{First Results
  in Robot Road-Following},'' in \emph{IJCAI}, 1985.

\bibitem{Furgale2010_JFR}
P.~Furgale and T.~D. Barfoot,
  ``\href{https://furgalep.github.io/sbib/furgale_jfr10.pdf}{Visual teach and
  repeat for long-range rover autonomy},'' \emph{JFR}, 2010.

\bibitem{Paull2017_ICRA}
L.~Paull, J.~Tani, H.~Ahn, J.~Alonso-Mora, L.~Carlone, M.~Cap, Y.~F. Chen,
  C.~Choi, J.~Dusek, Y.~Fang, \emph{et~al.},
  ``\href{http://michalcap.net/wp-content/papercite-data/pdf/paull_2017.pdf}{Duckietown:
  an open, inexpensive and flexible platform for autonomy education and
  research},'' in \emph{ICRA}, 2017.

\bibitem{Starship2020}
\BIBentryALTinterwordspacing
S.~Technologies. (2020) How starship delivery robots know where they are going.
  [Online]. Available:
  \url{https://medium.com/starshiptechnologies/how-starship-delivery-robots-know-where-they-are-going-c97d385a1015}
\BIBentrySTDinterwordspacing

\bibitem{How2008_CSM}
J.~P. How, B.~Behihke, A.~Frank, D.~Dale, and J.~Vian,
  ``\href{https://ieeexplore.ieee.org/document/4472379}{Real-time indoor
  autonomous vehicle test environment},'' \emph{IEEE Control Systems Magazine},
  2008.

\bibitem{Shen2011_ICRA}
S.~Shen, N.~Michael, and V.~Kumar,
  ``\href{https://ieeexplore.ieee.org/abstract/document/5980357}{Autonomous
  multi-floor indoor navigation with a computationally constrained MAV},'' in
  \emph{ICRA}, 2011.

\bibitem{Montemerlo2008_JFR}
M.~Montemerlo, J.~Becker, S.~Bhat, H.~Dahlkamp, D.~Dolgov, S.~Ettinger,
  D.~Haehnel, T.~Hilden, G.~Hoffmann, B.~Huhnke, \emph{et~al.},
  ``\href{http://robots.stanford.edu/papers/junior08.pdf}{Junior: The stanford
  entry in the urban challenge},'' \emph{JFR}, 2008.

\bibitem{Ort2018_ICRA}
T.~Ort, L.~Paull, and D.~Rus,
  ``\href{https://toyota.csail.mit.edu/sites/default/files/documents/papers/ICRA2018_AutonomousVehicleNavigationRuralEnvironment.pdf}{Autonomous
  vehicle navigation in rural environments without detailed prior maps},'' in
  \emph{ICRA}, 2018.

\bibitem{Fuentes2015_AIR}
J.~Fuentes-Pacheco, J.~Ruiz-Ascencio, and J.~M. Rend{\'o}n-Mancha,
  ``\href{https://link.springer.com/article/10.1007/s10462-012-9365-8}{Visual
  simultaneous localization and mapping: a survey},'' \emph{Artificial
  Intelligence Review}, 2015.

\bibitem{Waymo}
\BIBentryALTinterwordspacing
Waymo. [Online]. Available: \url{https://waymo.com}
\BIBentrySTDinterwordspacing

\bibitem{Pomerleau1989_NeurIPS}
D.~A. Pomerleau,
  ``\href{https://papers.nips.cc/paper/95-alvinn-an-autonomous-land-vehicle-in-a-neural-network.pdf}{ALVINN:
  An autonomous land vehicle in a neural network},'' in \emph{NeurIPS}, 1989.

\bibitem{Michels2005_ICML}
J.~Michels, A.~Saxena, and A.~Y. Ng,
  ``\href{http://ai.stanford.edu/~asaxena/rccar/ICML_ObstacleAvoidance.pdf}{High
  speed obstacle avoidance using monocular vision and reinforcement
  learning},'' in \emph{ICML}, 2005.

\bibitem{Fu2018_CVPR}
H.~Fu, M.~Gong, C.~Wang, K.~Batmanghelich, and D.~Tao,
  ``\href{https://arxiv.org/pdf/1806.02446.pdf}{Deep ordinal regression network
  for monocular depth estimation},'' in \emph{CVPR}, 2018.

\bibitem{Chang2018_CVPR}
J.-R. Chang and Y.-S. Chen,
  ``\href{https://arxiv.org/pdf/1803.08669.pdf}{Pyramid stereo matching
  network},'' in \emph{CVPR}, 2018.

\bibitem{Teichmann2018_IV}
M.~Teichmann, M.~Weber, M.~Zoellner, R.~Cipolla, and R.~Urtasun,
  ``\href{https://ieeexplore.ieee.org/abstract/document/8500504?casa_token=Y2O-CE_gtrEAAAAA:OKFwIRYVurGYBjmXmO-nU25rjNQ3xq1gs41pAJBNx0VTkqChHCzdokV35d9h4EqZ-0PJXBsXoMYo}{Multinet:
  Real-time joint semantic reasoning for autonomous driving},'' in \emph{IV},
  2018.

\bibitem{Wang2019_CVPR}
Y.~Wang, W.-L. Chao, D.~Garg, B.~Hariharan, M.~Campbell, and K.~Q. Weinberger,
  ``\href{https://arxiv.org/pdf/1812.07179.pdf}{Pseudo-lidar from visual depth
  estimation: Bridging the gap in 3d object detection for autonomous
  driving},'' in \emph{CVPR}, 2019.

\bibitem{Muller2006_NeurIPS}
U.~Muller, J.~Ben, E.~Cosatto, B.~Flepp, and Y.~L. Cun,
  ``\href{http://yann.lecun.com/exdb/publis/pdf/lecun-dave-05.pdf}{Off-road
  obstacle avoidance through end-to-end learning},'' in \emph{NeurIPS}, 2006.

\bibitem{Bojarski2016_arxiv}
M.~Bojarski, D.~Del~Testa, D.~Dworakowski, B.~Firner, B.~Flepp, P.~Goyal, L.~D.
  Jackel, M.~Monfort, U.~Muller, J.~Zhang, \emph{et~al.},
  ``\href{https://arxiv.org/abs/1604.07316}{End to end learning for
  self-driving cars},'' \emph{arXiv preprint arXiv:1604.07316}, 2016.

\bibitem{Pan2018_RSS}
Y.~Pan, C.-A. Cheng, K.~Saigol, K.~Lee, X.~Yan, E.~Theodorou, and B.~Boots,
  ``\href{https://arxiv.org/pdf/1709.07174.pdf}{Agile autonomous driving using
  end-to-end deep imitation learning},'' in \emph{RSS}, 2018.

\bibitem{Codevilla2018_ICRA}
F.~Codevilla, M.~Miiller, A.~L{\'o}pez, V.~Koltun, and A.~Dosovitskiy,
  ``\href{http://vladlen.info/papers/conditional-imitation.pdf}{End-to-end
  driving via conditional imitation learning},'' in \emph{ICRA}, 2018.

\bibitem{Ross2013_ICRA}
S.~Ross, N.~Melik-Barkhudarov, K.~S. Shankar, A.~Wendel, D.~Dey, J.~A. Bagnell,
  and M.~Hebert, ``\href{https://arxiv.org/pdf/1211.1690.pdf}{Learning
  monocular reactive uav control in cluttered natural environments},'' in
  \emph{ICRA}, 2013.

\bibitem{Giusti2015_RAL}
A.~Giusti, J.~Guzzi, D.~C. Cire{\c{s}}an, F.-L. He, J.~P. Rodr{\'\i}guez,
  F.~Fontana, M.~Faessler, C.~Forster, J.~Schmidhuber, G.~Di~Caro,
  \emph{et~al.}, ``\href{https://ieeexplore.ieee.org/document/7358076}{A
  machine learning approach to visual perception of forest trails for mobile
  robots},'' \emph{RA-L}, 2015.

\bibitem{Loquercio2018_RAL}
A.~Loquercio, A.~I. Maqueda, C.~R. Del-Blanco, and D.~Scaramuzza,
  ``\href{http://rpg.ifi.uzh.ch/docs/RAL18_Loquercio.pdf}{Dronet: Learning to
  fly by driving},'' \emph{RA-L}, 2018.

\bibitem{Sutton1998}
R.~S. Sutton, A.~G. Barto, \emph{et~al.},
  \emph{\href{http://www.incompleteideas.net/book/the-book-2nd.html}{Introduction
  to reinforcement learning}}, 1998.

\bibitem{Sadeghi2017_RSS}
F.~Sadeghi and S.~Levine, ``\href{https://arxiv.org/abs/1611.04201}{CAD2RL:
  Real single-image flight without a single real image},'' in \emph{RSS}, 2017.

\bibitem{Muller2018_CoRL}
M.~M{\"u}ller, A.~Dosovitskiy, B.~Ghanem, and V.~Koltun,
  ``\href{http://vladlen.info/papers/driving-policy-transfer.pdf}{Driving
  policy transfer via modularity and abstraction},'' in \emph{CoRL}, 2018.

\bibitem{Hirose2019_RAL}
N.~Hirose, F.~Xia, R.~Mart{\'\i}n-Mart{\'\i}n, A.~Sadeghian, and S.~Savarese,
  ``\href{https://arxiv.org/pdf/1903.02749.pdf}{Deep visual MPC-policy learning
  for navigation},'' \emph{RA-L}, 2019.

\bibitem{Riedmiller2007_FBIT}
M.~Riedmiller, M.~Montemerlo, and H.~Dahlkamp,
  ``\href{http://ml.informatik.uni-freiburg.de/former/_media/publications/riefbit07.pdf}{Learning
  to drive a real car in 20 minutes},'' in \emph{FBIT}, 2007.

\bibitem{Lipton2016_arxiv}
Z.~C. Lipton, K.~Azizzadenesheli, A.~Kumar, L.~Li, J.~Gao, and L.~Deng,
  ``\href{https://arxiv.org/pdf/1611.01211.pdf}{Combating reinforcement
  learning's sisyphean curse with intrinsic fear},'' \emph{arXiv preprint
  arXiv:1611.01211}, 2016.

\bibitem{Gandhi2017_IROS}
D.~Gandhi, L.~Pinto, and A.~Gupta,
  ``\href{https://arxiv.org/pdf/1704.05588.pdf}{Learning to fly by crashing},''
  in \emph{IROS}, 2017.

\bibitem{Kahn2017_arxiv}
G.~Kahn, A.~Villaflor, V.~Pong, P.~Abbeel, and S.~Levine,
  ``\href{https://arxiv.org/abs/1702.01182}{Uncertainty-aware reinforcement
  learning for collision avoidance},'' \emph{arXiv preprint arXiv:1702.01182},
  2017.

\bibitem{Richter2017_RSS}
C.~Richter and N.~Roy,
  ``\href{http://www.roboticsproceedings.org/rss13/p64.pdf}{Safe visual
  navigation via deep learning and novelty detection},'' in \emph{RSS}, 2017.

\bibitem{Saunders2018_AAMAS}
W.~Saunders, G.~Sastry, A.~Stuhlmueller, and O.~Evans,
  ``\href{https://arxiv.org/pdf/1707.05173.pdf}{Trial without error: Towards
  safe reinforcement learning via human intervention},'' in \emph{AAMAS}, 2018.

\bibitem{Kendall2019_ICRA}
A.~Kendall, J.~Hawke, D.~Janz, P.~Mazur, D.~Reda, J.-M. Allen, V.-D. Lam,
  A.~Bewley, and A.~Shah,
  ``\href{https://arxiv.org/pdf/1807.00412.pdf}{Learning to drive in a day},''
  in \emph{ICRA}, 2019.

\bibitem{Kahn2020_arxiv}
G.~Kahn, P.~Abbeel, and S.~Levine,
  ``\href{https://arxiv.org/abs/2002.05700}{BADGR: An autonomous
  self-supervised learning-based navigation system},'' \emph{arXiv preprint
  arXiv:2002.05700}, 2020.

\bibitem{Sandler2018_CVPR}
M.~Sandler, A.~Howard, M.~Zhu, A.~Zhmoginov, and L.-C. Chen,
  ``\href{https://arxiv.org/pdf/1801.04381.pdf}{MobileNetV2: Inverted residuals
  and linear bottlenecks},'' in \emph{CVPR}, 2018.

\bibitem{Hochreiter1997_neural}
S.~Hochreiter and J.~Schmidhuber,
  ``\href{https://www.bioinf.jku.at/publications/older/2604.pdf}{Long
  short-term memory},'' \emph{Neural computation}, 1997.

\bibitem{Camacho2013_springer}
E.~F. Camacho and C.~B. Alba, \emph{Model predictive control}, 2013.

\bibitem{Nagabandi2019_CoRL}
A.~Nagabandi, K.~Konoglie, S.~Levine, and V.~Kumar,
  ``\href{https://arxiv.org/pdf/1909.11652.pdf}{Deep Dynamics Models for
  Learning Dexterous Manipulation},'' in \emph{CoRL}, 2019.

\bibitem{Rubinstein2013_springer}
R.~Y. Rubinstein and D.~P. Kroese,
  \emph{\href{https://www.springer.com/gp/book/9780387212401}{The cross-entropy
  method: a unified approach to combinatorial optimization, Monte-Carlo
  simulation and machine learning}}.\hskip 1em plus 0.5em minus 0.4em\relax
  Springer Science \& Business Media, 2013.

\bibitem{Deng2009_CVPR}
J.~Deng, W.~Dong, R.~Socher, L.-J. Li, K.~Li, and L.~Fei-Fei,
  ``\href{http://www.image-net.org/papers/imagenet_cvpr09.pdf}{Imagenet: A
  large-scale hierarchical image database},'' in \emph{CVPR}, 2009.

\bibitem{Hessel2018_AAAI}
M.~Hessel, J.~Modayil, H.~Van~Hasselt, T.~Schaul, G.~Ostrovski, W.~Dabney,
  D.~Horgan, B.~Piot, M.~Azar, and D.~Silver,
  ``\href{https://arxiv.org/pdf/1710.02298.pdf}{Rainbow: Combining improvements
  in deep reinforcement learning},'' in \emph{AAAI}, 2018.

\bibitem{Kingma2014_ICLR}
D.~P. Kingma and M.~Welling,
  ``\href{https://arxiv.org/pdf/1312.6114.pdf}{Auto-encoding variational
  bayes},'' in \emph{ICLR}, 2014.

\bibitem{Lillicrap2016_ICLR}
T.~P. Lillicrap, J.~J. Hunt, A.~Pritzel, N.~Heess, T.~Erez, Y.~Tassa,
  D.~Silver, and D.~Wierstra,
  ``\href{https://arxiv.org/pdf/1509.02971.pdf}{Continuous control with deep
  reinforcement learning},'' in \emph{ICLR}, 2016.

\end{thebibliography}

\end{document}